\documentclass[journal]{IEEEtran}
\usepackage{amsmath,amssymb}
\usepackage{graphicx}
\usepackage{url}
\usepackage{color,colortbl}
\usepackage{xr}
\definecolor{cgood}{rgb}{0.55,0.85,0.55}
\definecolor{cok}{rgb}{1.0,0.9,0.55}
\definecolor{cmid}{rgb}{1.0,0.9,0.55}
\definecolor{cwor}{rgb}{1.0,0.72,0.45}
\definecolor{cbad}{rgb}{0.95,0.6,0.6}
\definecolor{cna}{rgb}{0.85,0.85,0.85}
\definecolor{car}{rgb}{0.93,0.93,0.93}

\newcommand{\cg}[1]{\cellcolor{cgood}#1}
\newcommand{\co}[1]{\cellcolor{cok}#1}
\newcommand{\cm}[1]{\cellcolor{cmid}#1}
\newcommand{\cw}[1]{\cellcolor{cwor}#1}
\newcommand{\cb}[1]{\cellcolor{cbad}#1}
\newcommand{\car}[1]{\cellcolor{car}#1}

\begin{document}

\title{Reproducible by Seed? Not on GPUs: Batch-Dependent Non-Determinism in Text-to-Image Models}

\author{Javier Conde, Xiuyuan Qi, Shanshan Liu, and Pedro Reviriego%
\thanks{J. Conde and P. Reviriego are with the Information Processing and Telecommunications Center, ETSI de Telecomunicaci\'on, Universidad Polit\'ecnica de Madrid, Madrid, Spain (e-mail: javier.conde.diaz@upm.es; pedro.reviriego@upm.es).}%
\thanks{X. Qi and S. Liu are with the University of Electronic Science and Technology of China, Chengdu, China (e-mail: xiuyuanqi@std.uestc.edu.cn; ssliu@uestc.edu.cn).}%
}


\maketitle

\begin{abstract}
In text to image AI models, fixing the prompt and the random seed is widely assumed to fix the image. We show that it does not on GPUs. We evaluate six text-to-image models (diffusion and autoregressive) under four numerical precisions and quantizations on three GPU types (RTX A6000, RTX 6000 Ada, and H100 NVL), across up to 100 prompts per set for all six models. For each configuration we compare, with the prompt and seed fixed, the image generated alone with the image generated inside a batch, and we also compare across GPUs. Every replay control reported here is bit-exact, so the variation is not random noise: the execution setting, mainly the batch composition and the GPU, decides the output. The effect is large and model dependent. Structural similarity between the single-image and the batched output falls to 0.580 for an autoregressive model in bf16 and to 0.551 with int8. For a second autoregressive model in bf16 it falls to 0.261 with sampled decoding and to 0.229 with greedy decoding, and for an fp8 diffusion model to 0.68. A diffusion model in bf16 stays close to the batch-1 output (0.96-0.98) but changes every image. In most cases the batched image is a different but equally valid image; in some, such as prompts with digits and counters, it is visibly worse. Cross-GPU differences are of the same order as the batch effect: for one autoregressive model in bf16 compared across two GPU architectures (RTX 6000 Ada versus RTX A6000), no image is bit-identical and the cross-GPU LPIPS (0.355 to 0.383) matches the batch effect within one GPU (0.347 to 0.381). The pattern is consistent with batch-dependent kernel execution: reduction orders change with tensor shapes, and 8-bit formats amplify small numerical differences into visible ones. The consequences are direct: seeds do not make results reproducible, benchmarks and image-quality comparisons that mix batch sizes or hardware can be biased, and safety audits and A/B tests on generated images need the serving setup reported.
\end{abstract}


\section{Introduction}

\IEEEPARstart{T}{ext-to-image} models are served at scale. To keep high efficiency in GPUs, inference engines group requests into batches, and to keep memory and cost down they run weights and activations in reduced precision, such as bf16, fp8, or int8. Users and benchmarks, on the other hand, assume that fixing the prompt and the seed fixes the image.

This assumption matters in several settings. Benchmarks of image quality compare models by their outputs. Audits and safety evaluations draw conclusions from generated images. A/B tests of serving stacks compare outputs before and after a change. All of these comparisons assume that, given the same prompt and seed, any difference in the output comes from the change being tested, not from the execution setting. A direct example is image editing. An editing model is asked to change one object in an image and to keep everything else the same \cite{hertz2023prompttoprompt,brooks2023instructpix2pix}. If the output depends on the execution setting, the same edit request can preserve the untouched region differently when the batch composition, the precision, or the hardware changes, so a region the user asked to keep can be altered by the serving setup rather than by the instruction. Preservation of the non-edited region is then an application-level consequence of execution-dependent generation, not only a numerical detail.

For large language models this assumption is known to fail. With greedy decoding and fixed seeds, the output of a request can change when the other requests in the batch change, because floating point addition is not associative and the reduction order depends on the batch shape \cite{atil2024nondeterminism,yuan2025understanding,he2025defeating,fu2026beyond}. Several fixes now exist, from batch-invariant kernels to scheduling schemes that isolate or pin the execution of a request \cite{he2025defeating,lmsys2025sglang,vllm2026batchinvariance,zhao2026corun,qi2026blocklock,gond2026llm42,xie2025repdl}.

For text-to-image models much less is known. Kim et al. showed that the same prompt and seed produce different images on different GPUs in Stable Diffusion, in the context of verification and consensus for generated content \cite{kim2023generative}. Fang et al. studied non-determinism in diffusion language models, a text-model family, under controlled factors \cite{fang2026dataset}. Practitioners have reported that generated images vary with batch size and across GPUs in widely used tools\footnote{diffusers issues \#2659 and \#6222, \url{https://github.com/huggingface/diffusers/issues/2659}, \url{https://github.com/huggingface/diffusers/issues/6222}; ComfyUI issue \#788, \url{https://github.com/comfyanonymous/ComfyUI/issues/788}; AUTOMATIC1111 issue \#10847, \url{https://github.com/AUTOMATIC1111/stable-diffusion-webui/issues/10847}.}. These reports are anecdotal; they do not measure the effect systematically.

In this paper we measure batch-dependent non-determinism in text-to-image generation. We fix the prompt, the seed, and the initial latent, and we compare the image generated alone with the image generated inside a batch, across models, precisions, batch sizes, and GPUs. To our knowledge, this is the first systematic evaluation of batch-size, quantization and hardware-induced non-determinism in text-to-image generation across diffusion and autoregressive models.

The contributions of this paper include:
\begin{enumerate}
\item A run protocol that separates execution-induced variation from random noise, with a replay control showing that identical runs are bit-exact and a staged design with nested prompt sets of 10, 25, and 100 prompts, so larger runs extend earlier results instead of replacing them.
\item Measurements on six models from the two main text-to-image families, diffusion and autoregressive, under four precisions and quantizations on three GPU types.
\item A quantification of the effect with four metrics (bit-identity, MAE, SSIM~\cite{wang2004ssim}, LPIPS~\cite{zhang2018lpips}), including worst cases in which the batched image is visibly worse, not only different.
\item An attribution of the effect to the batch, the precision, and the GPU. The autoregressive models show larger batch effects than every diffusion configuration we measured. We discuss the consequences for benchmarking, auditing, and serving.
\end{enumerate}

The rest of the paper is organized as follows. Section II reviews related work. Section III describes the models, hardware, prompts, protocol, and metrics. Section IV presents the results. Section V discusses the consequences and the limitations, and Section VI concludes.

\section{Related Work}

\subsection{Non-determinism in neural network execution}

Floating point addition is not associative, so the result of a parallel reduction depends on the order in which partial sums are combined. The impact of this property on the reproducibility of HPC and deep learning workloads is well documented \cite{shanmugavelu2024impacts}. Deep learning frameworks add their own sources of non-determinism, such as autotuned algorithm selection and atomic operations, which the framework documentation itself lists \cite{pytorch2026reproducibility,nvidia2023determinism}. Gundersen et al. review the sources of irreproducibility in machine learning more broadly \cite{gundersen2022sources}, and Summers and Dinneen show that in training, even one-bit changes in the initial parameters lead to vastly different final models \cite{summers2021nondeterminism}.

For LLM inference, Atil et al. measured how ``deterministic'' settings fail to be deterministic in practice \cite{atil2024nondeterminism}, and Yuan et al. traced the numerical sources of this behavior and proposed mitigations \cite{yuan2025understanding}. In our previous work we showed that token probabilities expose LLM nondeterminism even when the generated text does not change \cite{fu2026beyond}. These works cover text models; they do not measure image generation.

\subsection{Deterministic inference schemes}

Several schemes make LLM inference reproducible. Batch-invariant kernels constrain the reduction order of every operator so the result does not depend on the batch shape; this approach is described by Thinking Machines \cite{he2025defeating} and is implemented in vLLM \cite{vllm2026batchinvariance} and SGLang \cite{lmsys2025sglang}. RepDL provides bit-level reproducible training and inference \cite{xie2025repdl}. LLM-42 allows a fast non-reproducible path and verifies the result, rolling back when it is not canonical \cite{gond2026llm42}. CoRun isolates the prefill of each request and uses fixed-shape batched decode \cite{zhao2026corun}, and BlockLock pins the tensor geometry of the protected request so its numerical path does not depend on its companions \cite{qi2026blocklock}. GeulPU \cite{woo2026geulpu} is an open-source engine that targets bit-identical logits for GGUF LLMs. All of these target text models. For diffusion models, serving work such as SynerDiff studies continuous batching for speed, not reproducibility \cite{zhou2026synerdiff}. To our knowledge no equivalent fix has been evaluated for text-to-image models, and our results measure the problem such a fix would need to solve.

\subsection{Text-to-image models and quantization}

Text-to-image generation is dominated by two families. Diffusion models iteratively denoise a latent conditioned on the prompt, with architectures evolving from U-Nets, as in SDXL \cite{podell2024sdxl}, to multimodal diffusion transformers, as in SD3 \cite{esser2024scaling}, Qwen-Image \cite{wu2025qwenimage}, HiDream-I1 and HiDream-O1-Image \cite{cai2025hidream,hidream2026o1image}, and FLUX 3 Klein \cite{bfl2026flux3klein}. Autoregressive models tokenize the image and predict tokens sequentially, as in Janus-Pro \cite{chen2025januspro}, Emu3.5 \cite{cui2025emu35}, and Infinity \cite{han2025infinity}. In both families, serving at scale relies on reduced precision. Quantization of diffusion models is an active research area, with post-training schemes such as Q-Diffusion \cite{li2023qdiffusion} and low-bitwidth floating point formats \cite{chen2024lowbitwidth}. This work evaluates the interaction of these execution choices with reproducibility, which the quantization literature does not measure.

\subsection{Non-determinism in image generation}

The closest works to ours are Kim et al. \cite{kim2023generative} and Fang et al. \cite{fang2026dataset}. Kim et al. showed that identical prompts and seeds do not yield bit-identical images across GPUs in Stable Diffusion, and used this to design consensus and verification mechanisms. Their measurement is cross-GPU and single-model; they do not isolate batch size or precision as factors. Fang et al. evaluate non-determinism in diffusion language models under controlled factors and show that dataset-level metrics can hide it. Their models generate text, not images. Practitioner reports in the main inference tools describe outputs that vary with batch size and across GPUs, but without controlled measurements (see footnote in Section I). None of these works measures batch-size, quantization and hardware-induced non-determinism across the current text-to-image model families, which is the focus of this paper.

\section{Methodology}

\subsection{Models and hardware}

We evaluate six text-to-image models covering both families: SDXL base 1.0 \cite{podell2024sdxl}, Qwen-Image-2.1 \cite{wu2025qwenimage}, FLUX 3 Klein \cite{bfl2026flux3klein}, and HiDream-O1 \cite{hidream2026o1image} for diffusion, and Emu3.5 \cite{cui2025emu35} and Janus-Pro \cite{chen2025januspro} for autoregressive generation. Models run on three GPU types from three generations: the NVIDIA RTX A6000 (Ampere, no fp8 support), the RTX 6000 Ada (Ada Lovelace, fp8), and the H100 NVL (Hopper, fp8), the last both as a full device and as MIG 3g.47gb partitions. Table~\ref{tab:configs} lists the models and decoding configurations. Quantized variants in the main matrix use torchao: int8 is weight-only (Int8WeightOnly on the language-model linears for HiDream-O1) and fp8 uses the torchao float8 format. FLUX 3 Klein is configured with a batch ladder of 8, 16, and 32. The GPU time and peak memory of the runs are reported later in Appendix (Table~\ref{tab:resources}).

\begin{table*}[!t]
\centering
\caption{Models and decoding configurations.}
\label{tab:configs}
\begin{tabular}{lllllll}
\hline
Model & Family & Parameters & Precisions & Resolution & Steps & Guidance \\
\hline
SDXL base 1.0 & Diffusion (U-Net) & $\sim$3.5B & fp16, bf16 & 1024$\times$1024 & 30 & 7.5 \\
Qwen-Image-2.1 & Diffusion (MMDiT) & 7B & bf16, fp8 & \shortstack{512$\times$512\\768$\times$768} & 40 & 1.0 (off) \\
FLUX 3 Klein & Diffusion (rectified flow) & 5.6B & bf16, int8, fp8 & 512$\times$512 & 50 & 3.5 \\
HiDream-O1 & Diffusion & 8B & bf16, int8, fp8 & 512$\times$512 & 50 & 5.0 \\
Emu3.5 & Autoregressive & 34B & bf16, int8 & 512$\times$512 (32$\times$32 tokens) & -- & greedy \\
Janus-Pro-7B & Autoregressive & 7B & bf16 & 384$\times$384 (576 tokens) & -- & greedy, sampled \\
\hline
\end{tabular}
\end{table*}

\subsection{Prompts}

\begin{figure*}[!t]
\centering
\scriptsize\setlength{\tabcolsep}{2pt}
\begin{tabular}{@{}lcccccc@{}}
 & \begin{tabular}[b]{@{}c@{}}Prompt 00\\(bullseye)\end{tabular} & \begin{tabular}[b]{@{}c@{}}Prompt 01\\(colored rings)\end{tabular} & \begin{tabular}[b]{@{}c@{}}Prompt 04\\(red brick wall)\end{tabular} & \begin{tabular}[b]{@{}c@{}}Prompt 07\\(herringbone bricks)\end{tabular} & \begin{tabular}[b]{@{}c@{}}Prompt 08\\(pen crosshatch)\end{tabular} & \begin{tabular}[b]{@{}c@{}}Prompt 09\\(fine grid hatch)\end{tabular} \\[1pt]
SDXL & \includegraphics[width=0.130\textwidth]{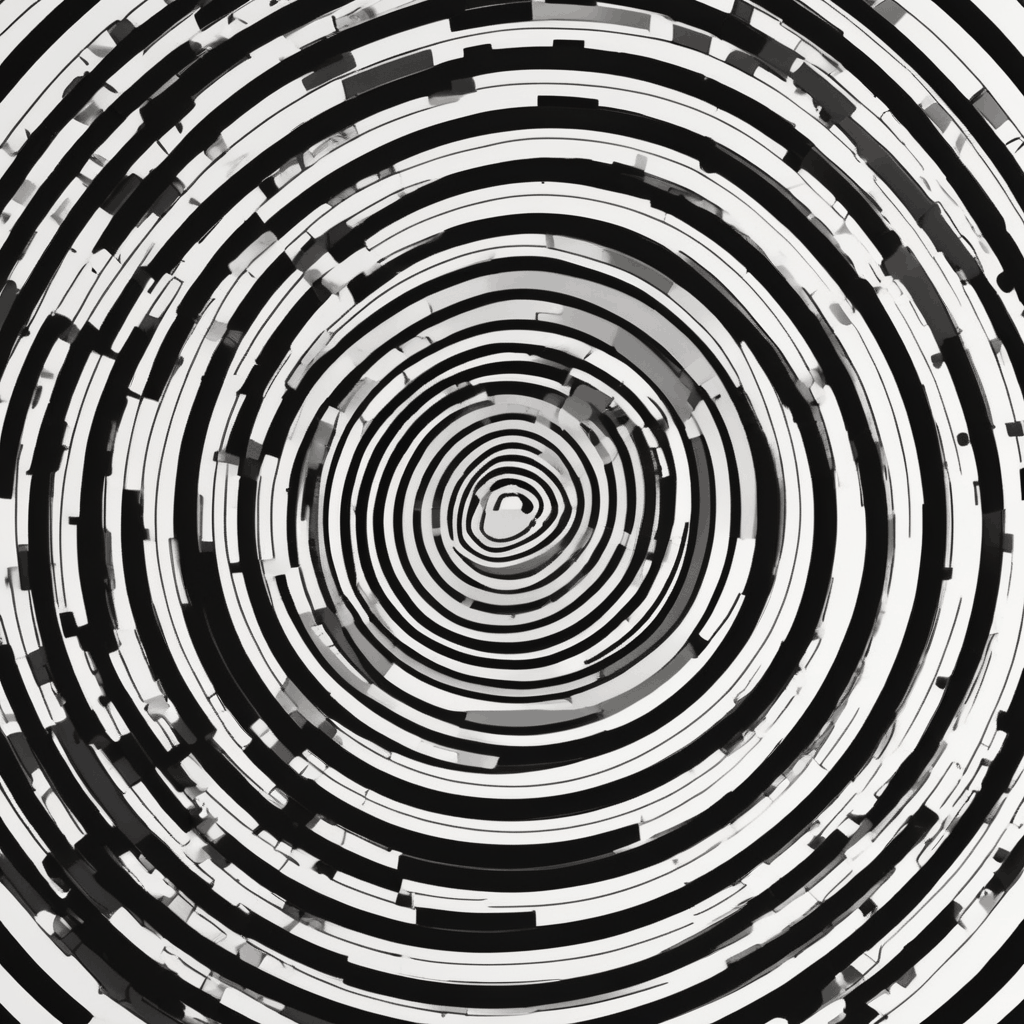} & \includegraphics[width=0.130\textwidth]{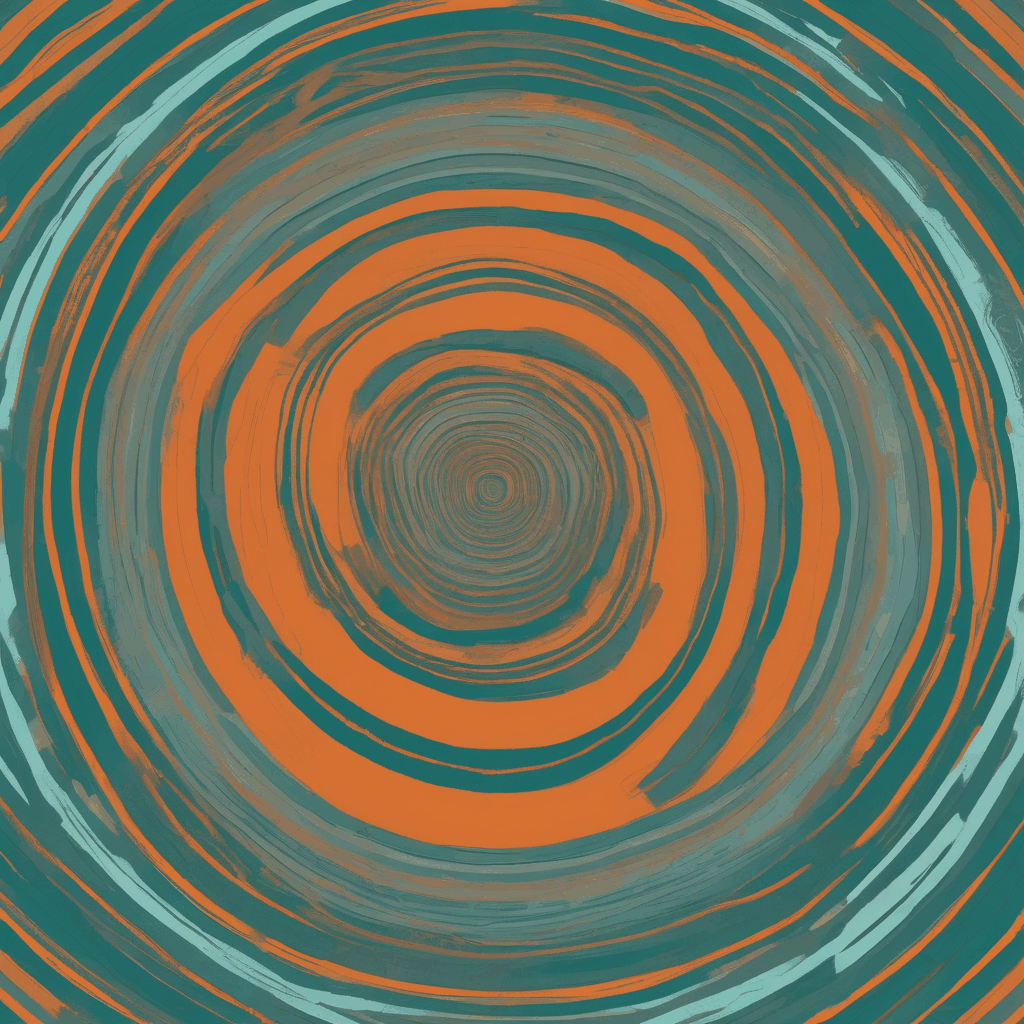} & \includegraphics[width=0.130\textwidth]{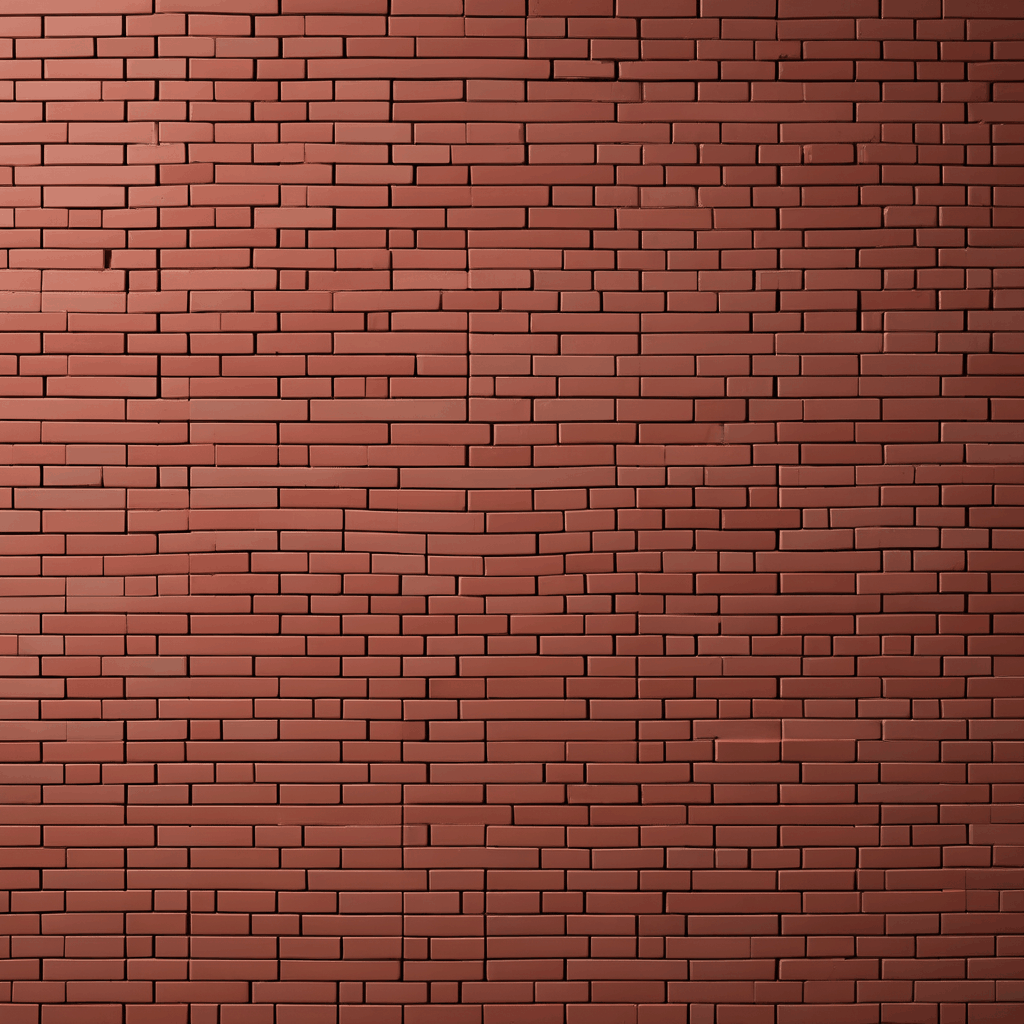} & \includegraphics[width=0.130\textwidth]{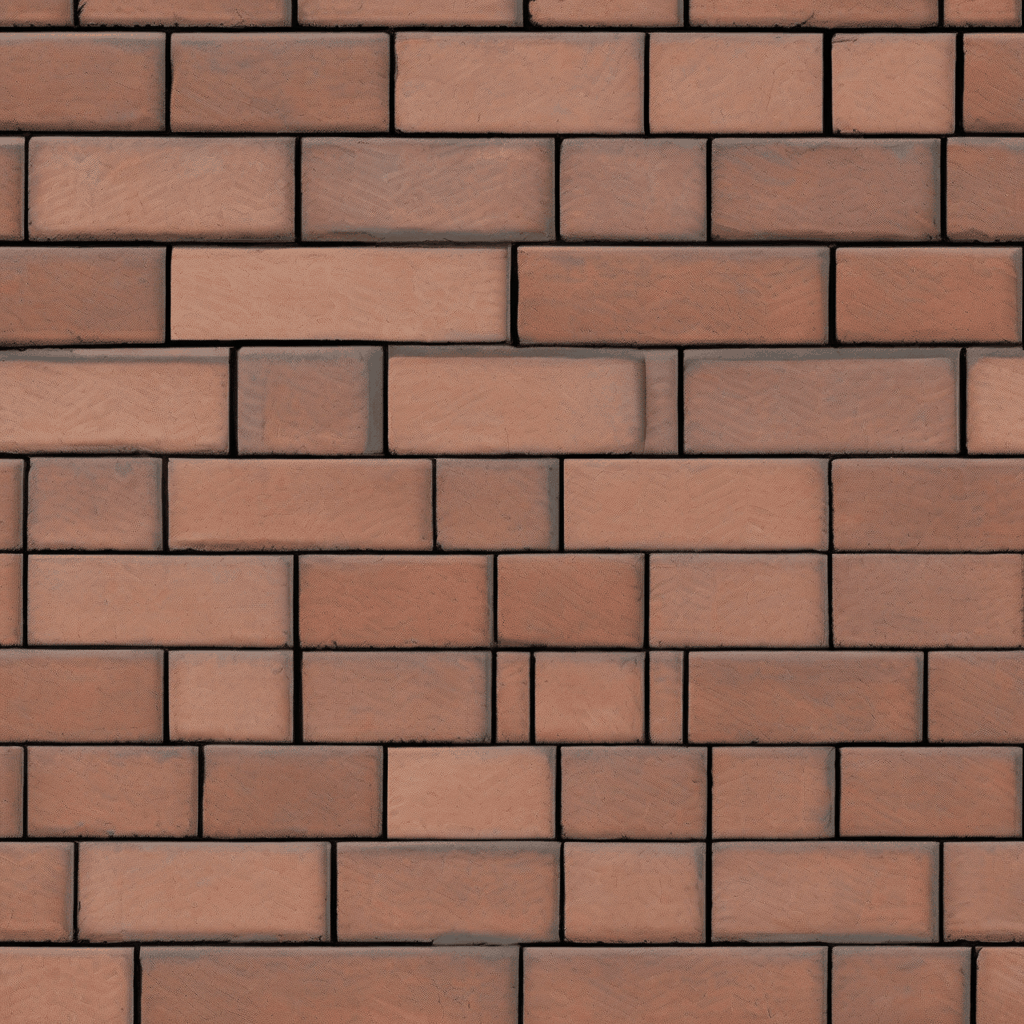} & \includegraphics[width=0.130\textwidth]{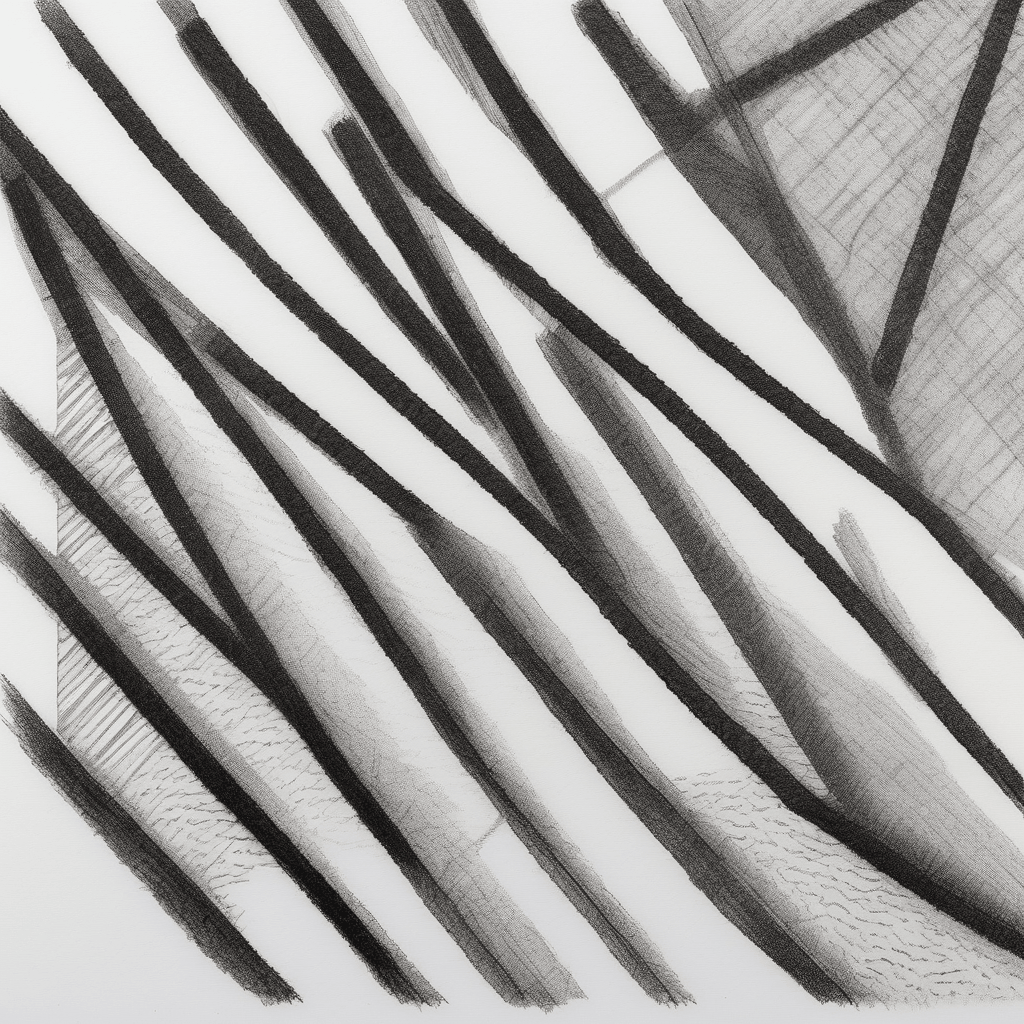} & \includegraphics[width=0.130\textwidth]{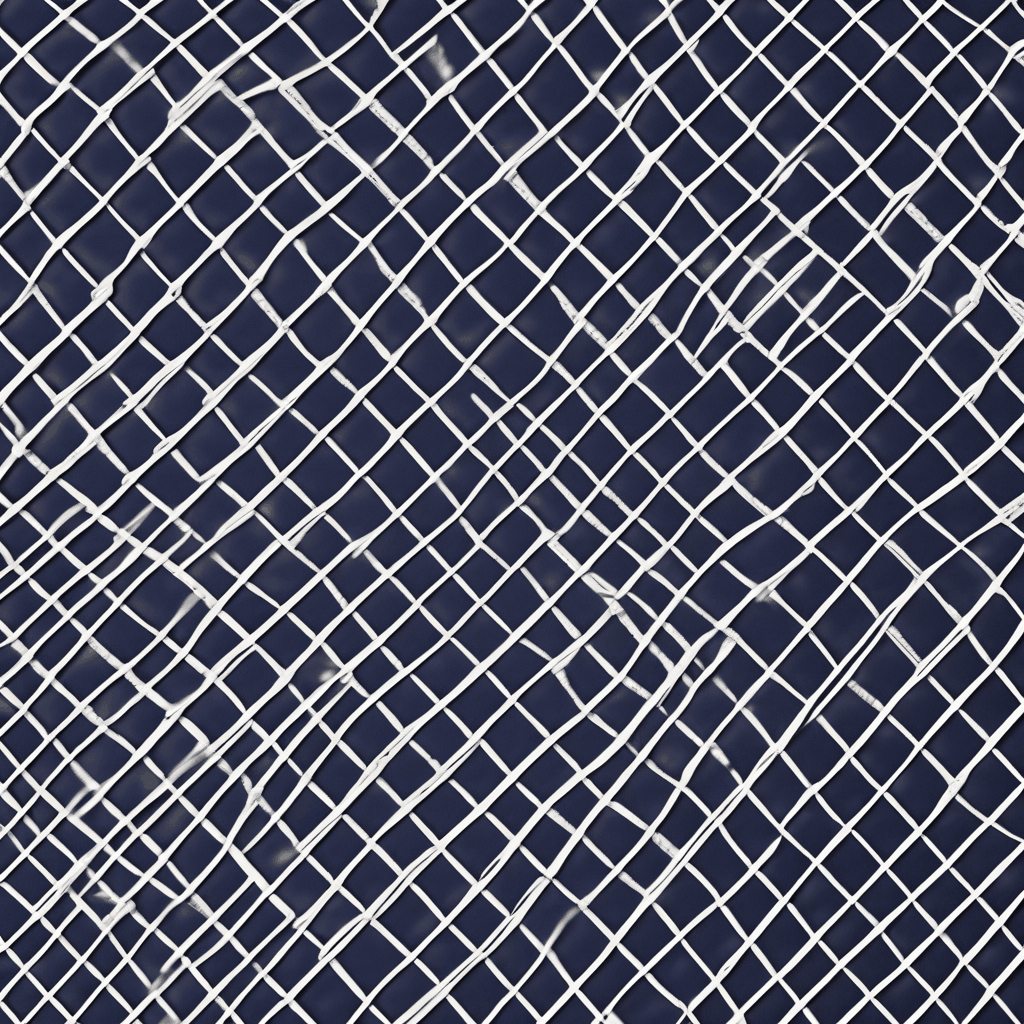} \\[1pt]
Qwen-Image-2.1 & \includegraphics[width=0.130\textwidth]{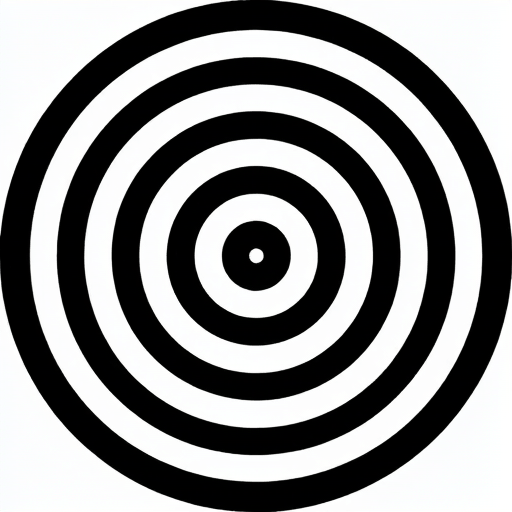} & \includegraphics[width=0.130\textwidth]{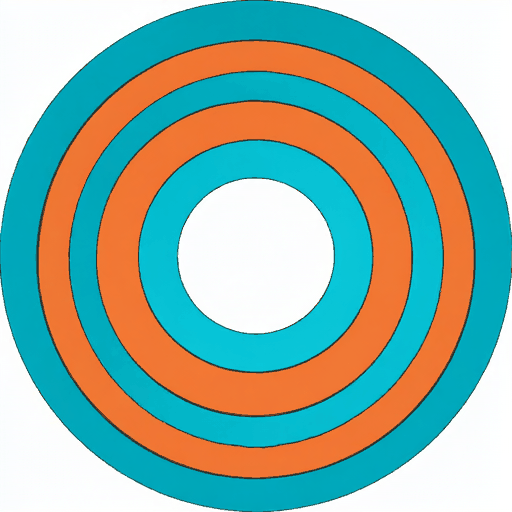} & \includegraphics[width=0.130\textwidth]{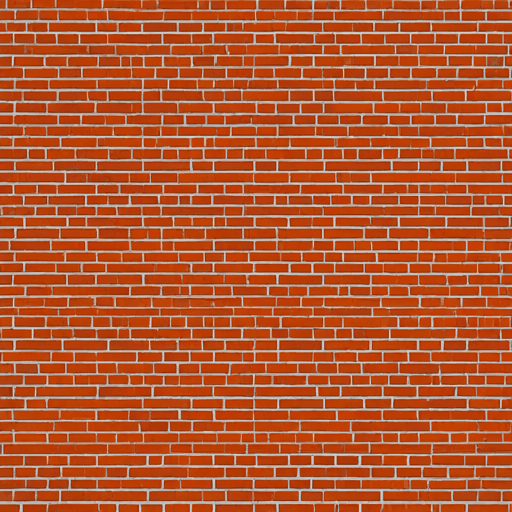} & \includegraphics[width=0.130\textwidth]{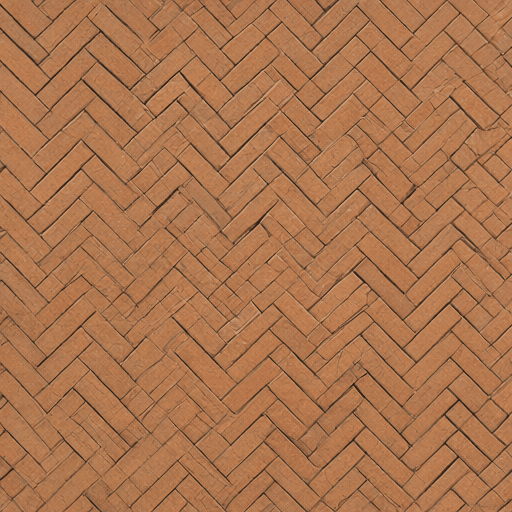} & \includegraphics[width=0.130\textwidth]{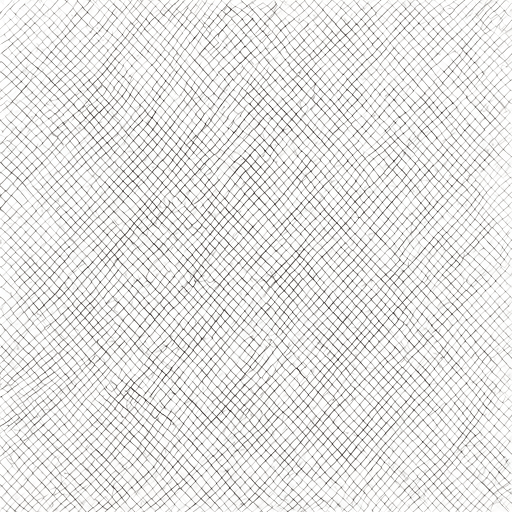} & \includegraphics[width=0.130\textwidth]{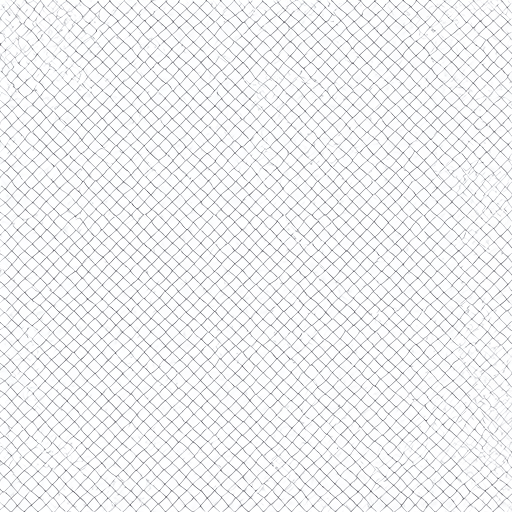} \\[1pt]
FLUX 3 Klein & \includegraphics[width=0.130\textwidth]{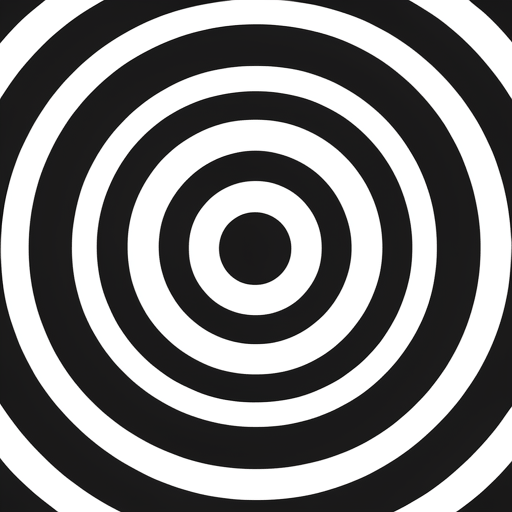} & \includegraphics[width=0.130\textwidth]{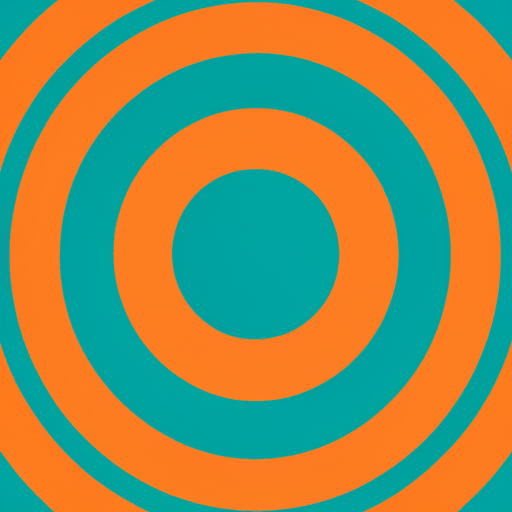} & \includegraphics[width=0.130\textwidth]{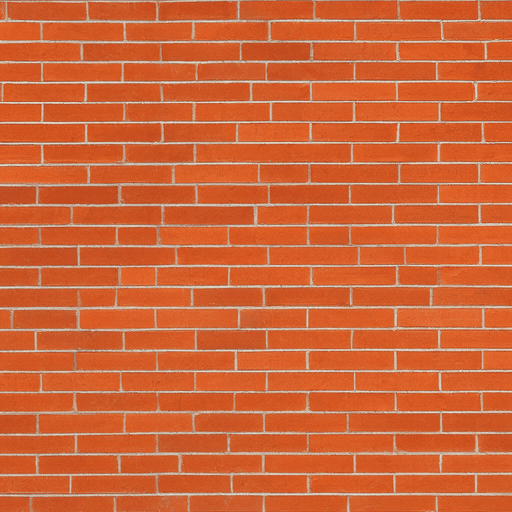} & \includegraphics[width=0.130\textwidth]{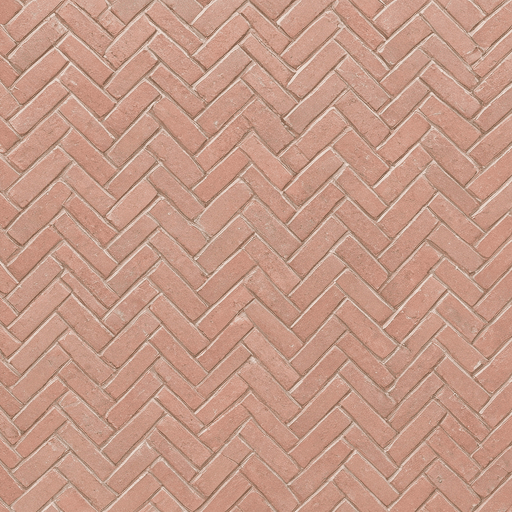} & \includegraphics[width=0.130\textwidth]{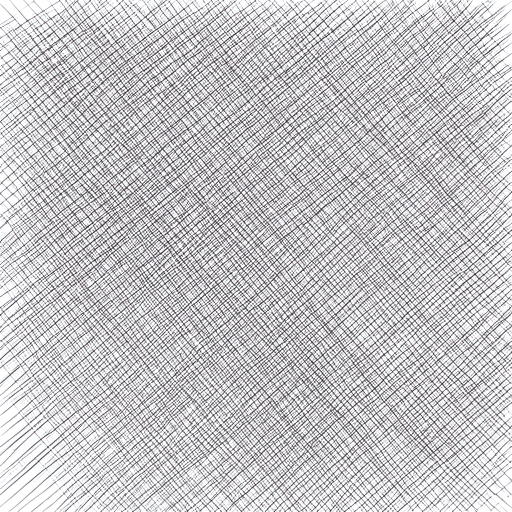} & \includegraphics[width=0.130\textwidth]{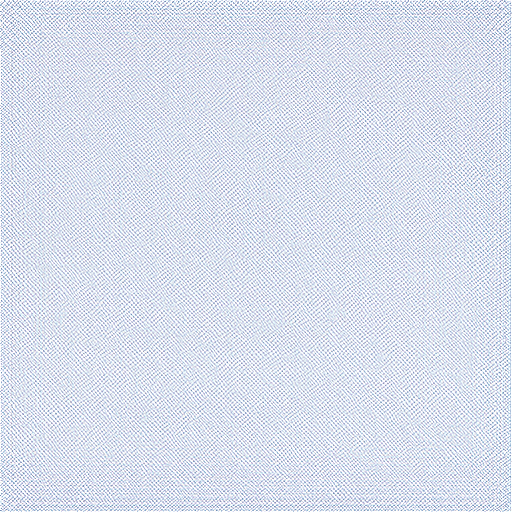} \\[1pt]
HiDream-O1 & \includegraphics[width=0.130\textwidth]{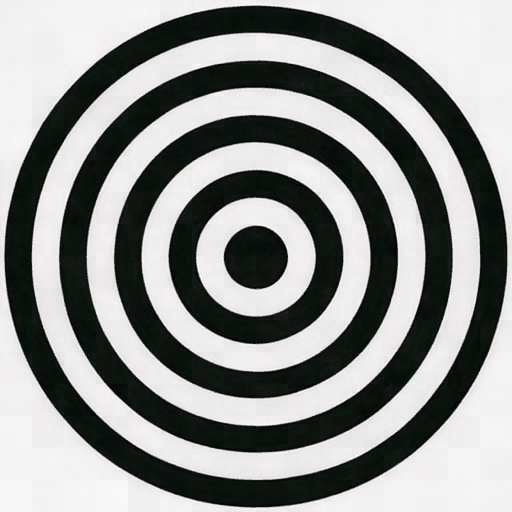} & \includegraphics[width=0.130\textwidth]{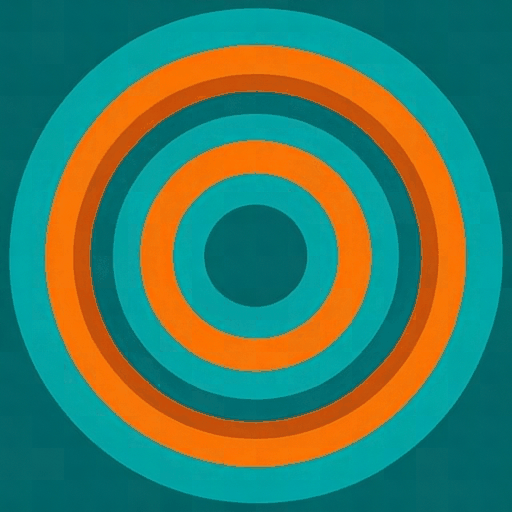} & \includegraphics[width=0.130\textwidth]{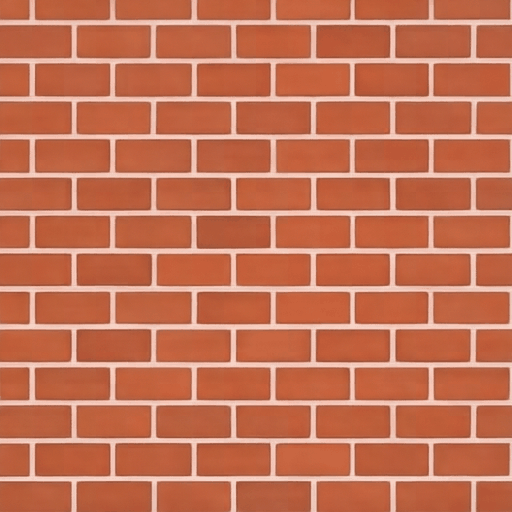} & \includegraphics[width=0.130\textwidth]{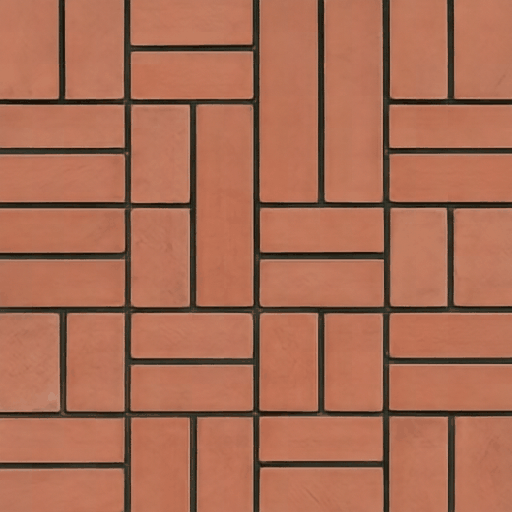} & \includegraphics[width=0.130\textwidth]{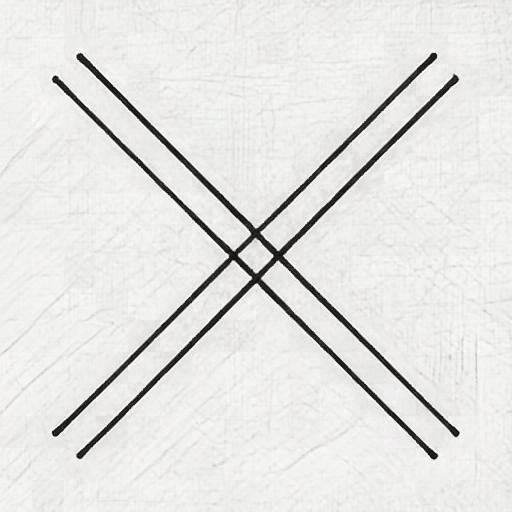} & \includegraphics[width=0.130\textwidth]{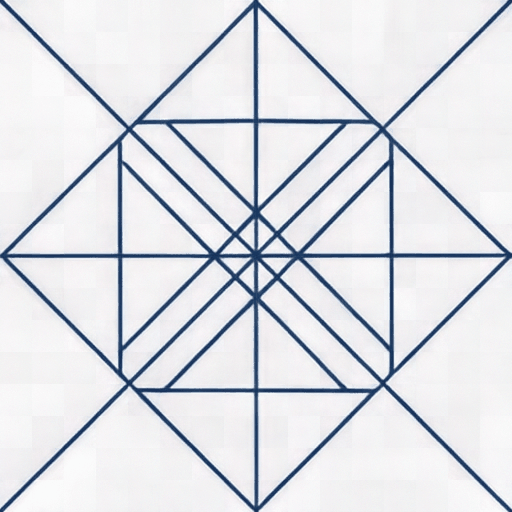} \\[1pt]
Emu3.5 & \includegraphics[width=0.130\textwidth]{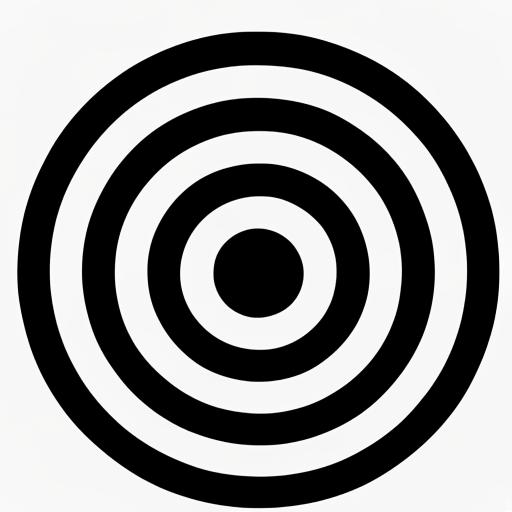} & \includegraphics[width=0.130\textwidth]{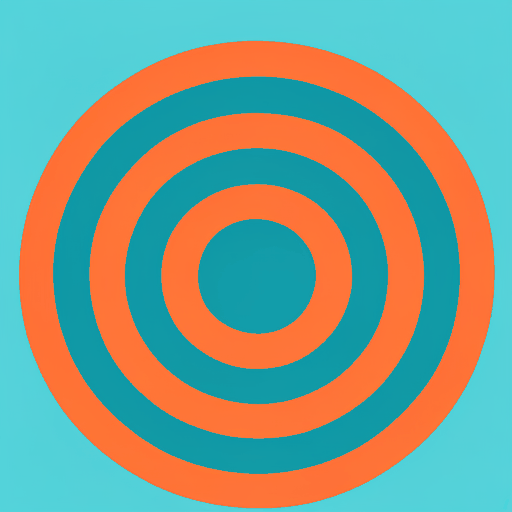} & \includegraphics[width=0.130\textwidth]{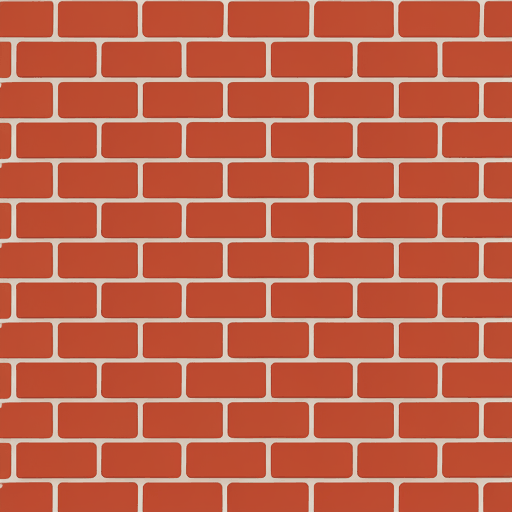} & \includegraphics[width=0.130\textwidth]{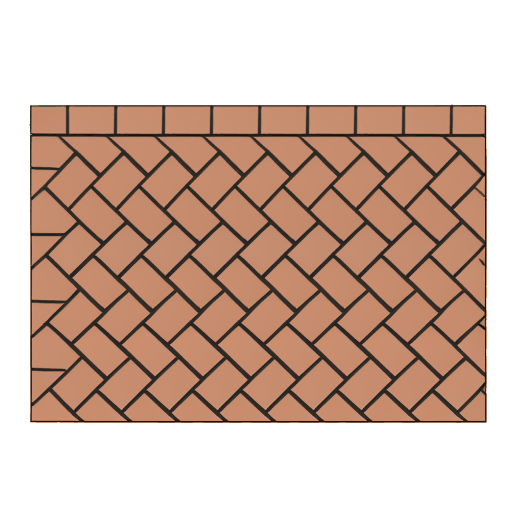} & \includegraphics[width=0.130\textwidth]{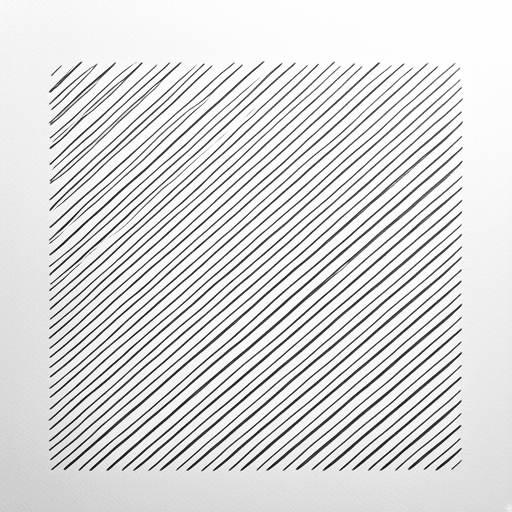} & \includegraphics[width=0.130\textwidth]{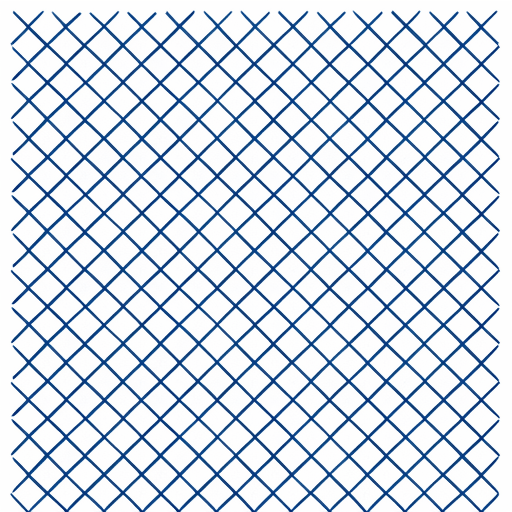} \\[1pt]
Janus-Pro-7B & \includegraphics[width=0.130\textwidth]{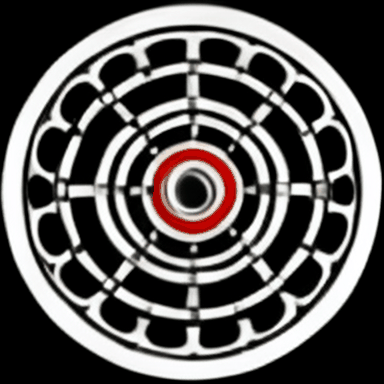} & \includegraphics[width=0.130\textwidth]{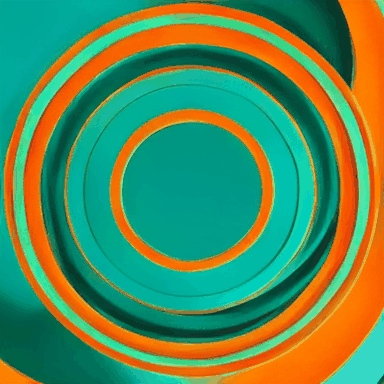} & \includegraphics[width=0.130\textwidth]{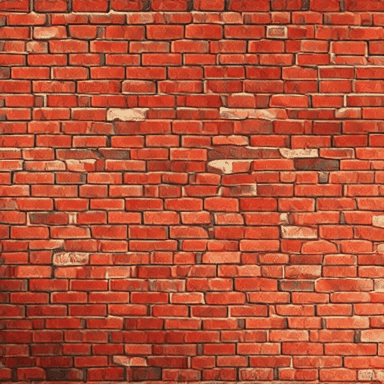} & \includegraphics[width=0.130\textwidth]{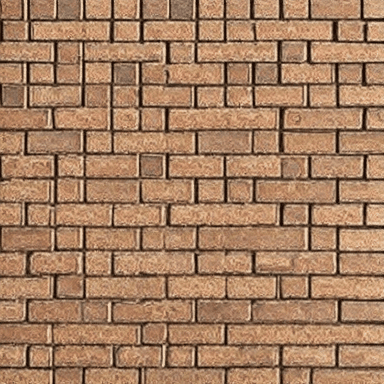} & \includegraphics[width=0.130\textwidth]{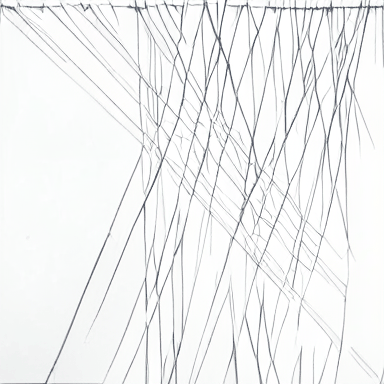} & \includegraphics[width=0.130\textwidth]{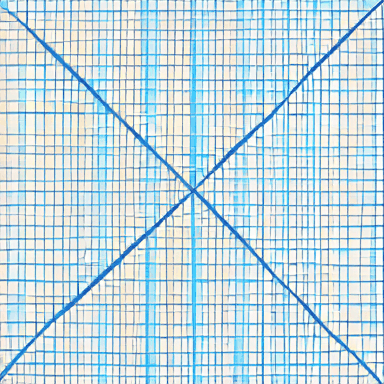} \\[1pt]
\end{tabular}
\caption{Batch-1 reference images (arm A, first run) of one run per model for six example prompts (columns: prompt 00 bullseye, prompt 01 colored rings, prompt 04 red brick wall, prompt 07 herringbone bricks, prompt 08 pen crosshatch, prompt 09 fine grid hatch), chosen without metric-based selection. SDXL runs in fp16 at 1024 px, Janus-Pro-7B at 384 px native, and the other models at 512 px. The Janus-Pro-7B prompts are wrapped in "A detailed image of ...".}
\label{fig:promptgrid}
\end{figure*}

The prompts are organized in $S$ sets of $P$ prompts, with one base seed per set and different seeds across sets. SDXL, FLUX 3 Klein, and HiDream-O1 use $S = 3$ sets and Qwen-Image-2.1 uses $S = 2$. The stages are nested: $P = 10$, 25, and 100 prompts per set. The autoregressive models use $P = 10$ and $S = 1$ for the main comparisons, with the same prompts for Emu3.5 and Janus-Pro. Both also ran 100-prompt sets (Emu3.5 in bf16 and int8 on the H100 NVL; Janus-Pro-7B in bf16 on the RTX 6000 Ada and the RTX A6000). The prompts cover several families: high-frequency periodic patterns (checkerboards, gratings, and angled stripes), geometric and organic scenes, colored patterns, and prompts with digits, text, and counters, such as scoreboards and seven-segment displays. The patterns sit near the resolution limit of the models, where small numerical changes are most likely to become visible. The text-bearing prompts probe semantic content that is either right or wrong. Figure~\ref{fig:promptgrid} shows the batch-1 reference image of each model for six example prompts.

\subsection{Run protocol}

For each model, precision, and prompt we generate the image alone and inside a batch, plus a replay check. The solo arms A0 and A1 each generate every prompt at batch size 1, and A0 (r0) is the reference for all comparisons. The batched arm B generates the image inside batches of 8 prompts with rotating membership and shuffled positions. The replay arm repeats a run verbatim to test run-to-run determinism: arm C for SDXL, Qwen-Image-2.1, and HiDream-O1, which repeats the round-0 batch, and arm R for FLUX 3 Klein, Emu3.5, and Janus-Pro-7B, which replays arm B (B with $n = 64$ at stage 10 for FLUX 3 Klein). The autoregressive models decode with greedy argmax, so there is no sampling seed (Janus-Pro also ran a sampled arm, $T = 1.0$, cfg 5.0). Their batched arm B uses one cell of 8 prompts and one of 2. Emu3.5 runs 10 images in each of arms A0, A1, B, and R at stage 10 and 100 per arm at stage 100. Janus-Pro-7B runs the same four arms in each of its two decoding modes: 80 images per run, 40 per mode, at stage 10, and 800 per run at 100 prompts. For the diffusion models, each cell is repeated 2 to 3 times with identical configuration. Table~\ref{tab:protocol} gives the protocol sizes per model. The protocol is parameterized by $P$, $S$, the seeds, and the batch cell, so a larger run is a configuration change, not a new protocol. The full study runs in three nested stages of $P = 10$, 25, and 100 prompts per set, with one seed per set. Each stage extends the previous one, so the results of each stage remain valid subsets of the final run. For SDXL the stage-100 runs cover one set of 100 prompts ($S = 1$ at that stage). Appendix~\ref{app:schedule} describes the run times and memory use.

\begin{table}[!t]
\centering
\caption{Protocol sizes per model and stage: prompts per set ($P$), sets ($S$), and images per arm. A0/A1: the two solo arms; B: batched; C or R: the replay (C for SDXL, Qwen-Image-2.1, and HiDream-O1, R for FLUX 3 Klein, Emu3.5, and Janus-Pro-7B). Qwen-Image-2.1's arm B is 64 at every stage (8 rounds of 8). Janus-Pro-7B sizes are per decoding mode. Totals per run (generated PNGs) are in Appendix~\ref{app:expdetails}.}
\label{tab:protocol}
\scriptsize\setlength{\tabcolsep}{1pt}
\begin{tabular}{lllllll}
\hline
Model & Run & $P$ & $S$ & A0/A1 & B & C/R \\
\hline
SDXL & s10/s25 & 10/25 & 3 & 10/10, 25/25 & 256/632 & 16 \\
SDXL (bf16) & s100 & 100 & 1 & 100/100 & 2504 & 16 \\
Qwen-Image-2.1 & s10 & 10 & 2 & 10/10 & 64 & 8 \\
Qwen-Image-2.1 & s25 & 25 & 2 & 25/25 & 64 & 8 \\
Qwen-Image-2.1 & s100 & 100 & 2 & 100/100 & 64 & 8 \\
Qwen-Image-2.1 & s100, B24 & 100 & 2 & 100/100 & 192 & 24 \\
FLUX 3 Klein & s10 & 10 & 3 & 10/10 & 64 & 64 \\
FLUX 3 Klein & s25 & 25 & 3 & 25/25 & 96 & 96 \\
FLUX 3 Klein & s100 & 100 & 3 & 100/100 & 344 & 344 \\
HiDream-O1 & s10 & 10 & 3 & 10/10 & 128 & 8 \\
HiDream-O1 & s25 & 25 & 3 & 25/25 & 128 & 8 \\
HiDream-O1 & s100, B8-B32 & 100 & 3 & 100/100 & 128/256/384/512 & 8/16/24/32 \\
Emu3.5 & s10 & 10 & 1 & 10/10 & 10 & 10 \\
Emu3.5 (bf16) & s100 & 100 & 1 & 100/100 & 100 & 100 \\
Emu3.5 (int8) & s100 & 100 & 1 & 100/100 & 100 & 100 \\
Janus-Pro-7B & s10 & 10 & 1 & 10/10 & 10 & 10 \\
Janus-Pro-7B (bf16) & p100 & 100 & 1 & 100/100 & 100 & 100 \\
\hline
\end{tabular}
\end{table}

\subsection{Metrics}

We report four metrics between each batched image and its batch-1 reference: bit-identity, MAE, SSIM, and LPIPS (Table~\ref{tab:metrics}). For bit-identical pairs the metrics are exactly zero and are reported without running the networks. All images are decoded to 8-bit RGB before comparison; comparisons are read-only and offline. Qwen-Image-2.1 PNGs carry an alpha channel, which is ignored for SSIM, MAE, and LPIPS. The SDXL LPIPS values and the Janus-Pro-7B cross-GPU LPIPS values are computed on CPU with the isolated LPIPS package (AlexNet backbone, linear calibration), a different analysis stack from the GPU-side pipeline; the LPIPS weights are byte-identical in both.

\begin{table}[!t]
\centering
\caption{The four metrics between each batched image and its batch-1 reference.}
\label{tab:metrics}
\footnotesize\setlength{\tabcolsep}{4pt}
\begin{tabular}{lp{5.8cm}}
\hline
Metric & What it measures \\
\hline
Bit-identity & Fraction of bit-identical images, where bit-identical means the decoded pixel arrays are exactly equal \\
MAE & Mean absolute error over pixels, on a 0 to 255 scale \\
SSIM & Structural similarity index~\cite{wang2004ssim}, computed with scikit-image 0.26.0 on RGB images at native resolution with data range 255, averaged over channels \\
LPIPS & Perceptual distance~\cite{zhang2018lpips} with the AlexNet backbone, using lpips 0.1.4 with inputs scaled to $[-1, 1]$ \\
\hline
\end{tabular}
\end{table}

\section{Results}

This section presents the empirical results. Raw generated images (A0, A1, B, and C/R arms) together with an interactive viewer for inspecting matched image pairs are released on Zenodo\footnote{https://doi.org/10.5281/zenodo.24444740}.

\subsection{Replay control}

\textbf{Replay.} Across the replay controls reported here, repeating the exact same run, same prompt, seed, batch composition, and process setup, produces bit-identical outputs for the models and precisions tested. The variation we report below is therefore not random noise. It is induced by the execution setting. As a concrete case, the three fp8 runs of HiDream-O1 on set 3 reproduce identically. A repeated Emu3.5 int8 run also reproduces the earlier run exactly: solo and batched outputs are both bit-identical on replay, 10 of 10 prompts each. Table~\ref{tab:replay} summarizes the replay control. The control also holds at the largest scale. For SDXL, the batch-1 repeats are bit-exact in both stage runs, and all 16 replay images differ from the solo reference in each run (16 of 16). At stage 100 both solo repeats are bit-exact for all 100 prompts (200 of 200 pairs), and both batch replays reproduce the batch exactly (16 of 16) while differing from solo (16 of 16). Emu3.5 behaves the same way at stage 100: a bf16 run on the H100 NVL generates 100 images per arm (A0, A1, B, R), all A0/A1 and B/R pairs are PNG-exact, all 400 decode-control replays are pixel-exact, and the int8 stage-100 run replays exactly as well. The stage-100 batch-versus-solo differences are large even though every replay is exact. In bf16 and int8 order: mean LPIPS 0.296 and 0.307 (maxima 0.673 and 0.804), mean SSIM 0.598 and 0.597 (minima 0.008 and 0.013, maxima 0.987 and 0.994), and mean MAE 34.571 and 32.849 (minima 2.028 and 0.662, maxima 115.465 and 113.596), with 0 of 100 pairs exact in both precisions.

\begin{table}[!t]
\centering
\caption{Replay control: identical reruns of the same run. Green: the rerun reproduces the original run bit-exactly. Light gray model cells mark the autoregressive models (Emu3.5, Janus-Pro-7B).}
\label{tab:replay}
\footnotesize\setlength{\tabcolsep}{4pt}
\begin{tabular}{llll}
\hline
Model & Precision & Batch-1 rerun & Batch replay \\
\hline
SDXL & fp16, bf16\footnotemark & \cg{yes} & \cg{yes} \\
Qwen-Image-2.1 & bf16, fp8 & \cg{yes} & \cg{yes} \\
FLUX 3 Klein & bf16, int8, fp8 & \cg{yes} & \cg{yes} \\
HiDream-O1 & bf16, int8 & \cg{yes} & \cg{yes} \\
HiDream-O1 & fp8 & \cg{yes (3 runs)} & \cg{yes} \\
\car{Emu3.5} & bf16, int8 & \cg{yes} & \cg{yes} \\
\car{Janus-Pro-7B} & bf16 & \cg{yes} & \cg{yes} \\
\hline
\end{tabular}
\end{table}
\footnotetext{SDXL fp16 and bf16 are from the dtype probe: 8 prompts, batch 8 versus batch 1.}

\subsection{Batch versus single image}

\textbf{Overview.} Table~\ref{tab:main} summarizes the comparison between batched and single-image generation for each model and precision. The table retains the measured configurations; cells without a measured metric remain marked as not measured. New GPU-specific runs are reported separately rather than substituted into another GPU row.

\begin{table}[!t]
\centering
\caption{Batch versus single-image generation with the same prompt and seed. SSIM against the batch-1 reference (1.0 is identical). Shading: green = identical or near-identical, yellow = sub-visible or small visible, orange = visible structural change, red = large structural change, dark gray = not measured. Light gray non-result cells mark the autoregressive models (Emu3.5, Janus-Pro-7B). The SDXL MAE is logged by the runner.}
\label{tab:main}
\scriptsize\setlength{\tabcolsep}{2pt}
\resizebox{\columnwidth}{!}{%
\begin{tabular}{lllllll}
\hline
Model & Prec. & GPU & Bit-id. & MAE & SSIM & LPIPS \\
\hline
SDXL & fp16 & A6000 & \cb{0.0} & \co{3.308} & \co{0.951} & \co{0.026} \\
SDXL & bf16 & A6000 & \cb{0.0} & \co{8.230} & \cm{0.878} & \cm{0.072} \\
SDXL (s10) & fp16 & A6000 & \cb{0.0} & \co{1.85} & \co{0.970} & \cm{0.013} \\
SDXL (s25) & fp16 & A6000 & \cb{0.0} & \co{2.164} & \co{0.973} & \cm{0.013} \\
Qwen-Image-2.1 & fp16 & H100 NVL & \cb{0.000} & \cm{6.789} & \co{0.878} & \cm{0.063} \\
Qwen-Image-2.1 & bf16 & H100 & \cb{0.0} & \cm{7.476} & \cm{0.873} & \cm{0.068} \\
Qwen-Image-2.1 (s10) & bf16 & H100 & \cb{0.0} & \cm{5.54} & \co{0.894} & \cm{0.047} \\
Qwen-Image-2.1 (s10) & fp8 & H100 & \cb{0.0} & \cm{7.75} & \co{0.811} & \cm{0.044} \\
Qwen-Image-2.1 (s25) & bf16 & H100 & \cb{0.0} & \cm{7.98} & \co{0.850} & \cm{0.068} \\
Qwen-Image-2.1 (s25) & fp8 & H100 & \cb{0.0} & \cm{9.72} & \co{0.820} & \cm{0.071} \\
Qwen-Image-2.1 (s100) & bf16 & H100 & \cb{0.0} & \cm{8.91} & \co{0.875} & \cm{0.088} \\
Qwen-Image-2.1 (s100) & fp8 & H100 & \cb{0.0} & \cm{10.00} & \co{0.863} & \cm{0.092} \\
FLUX 3 Klein & bf16 & H100-M & \cb{0.0} & \cg{0.173} & \cg{0.999} & \cg{$\approx$0} \\
FLUX 3 Klein (s10) & bf16 & H100-M & \cb{0.0} & \cg{0.359} & \cg{0.998} & \cg{$\approx$0} \\
FLUX 3 Klein (s10) & int8 & H100-M & \cb{0.0} & \cg{0.356} & \cg{0.998} & \cg{$\approx$0} \\
FLUX 3 Klein (s10) & fp8 & H100-M & \cb{0.0} & \cw{17.33} & \co{0.829} & \cm{0.074} \\
FLUX 3 Klein (s100) & fp8 & H100-M & \cb{0.0} & \cm{8.86} & \co{0.893} & \cm{0.045} \\
FLUX 3 Klein (s100) & bf16 & H100-M & \cb{0.0} & \cg{0.127} & \cg{0.999} & \cg{$\approx$0} \\
FLUX 3 Klein (s100) & int8 & H100-M & \cb{0.0} & \cg{0.117} & \cg{0.999} & \cg{$\approx$0} \\
HiDream-O1 & bf16 & H100-M & \cg{1.00} & \cg{0.0} & \cg{1.000} & \cg{0.0} \\
HiDream-O1 & int8 & H100-M & \cg{1.00} & \cg{0.0} & \cg{1.000} & \cg{0.0} \\
HiDream-O1 & fp8 & H100-M & \cb{0.0} & \cw{15.45} & \cw{0.748} & \cm{0.109} \\
HiDream-O1 (s25) & fp8 & H100-M & \cb{0.0} & \cw{17.85} & \cw{0.731} & \cm{0.159} \\
HiDream-O1 (s100) & bf16 & H100-M & \cg{1.00} & \cg{0.0} & \cg{1.000} & \cg{0.0} \\
HiDream-O1 (s100) & int8 & H100-M & \cg{1.00} & \cg{0.0} & \cg{1.000} & \cg{0.0} \\
HiDream-O1 (s100) & fp8 & H100-M & \cb{0.0} & \cw{19.56} & \cw{0.686} & \cm{0.176} \\
\car{Emu3.5} & \car{bf16} & \car{H100} & \cb{0.0} & \cb{30.16} & \cw{0.580} & \cm{0.168} \\
\car{Emu3.5} & \car{int8} & \car{H100} & \cb{0.0} & \cb{31.78} & \cw{0.551} & \cw{0.229} \\
\car{Janus-Pro-7B (samp.)} & \car{bf16} & \car{Ada} & \cb{0.0} & \cb{46.64} & \cb{0.261} & \cw{0.296} \\
\car{Janus-Pro-7B (greedy)} & \car{bf16} & \car{Ada} & \cb{0.0} & \cb{54.77} & \cb{0.229} & \cw{0.319} \\
\hline
\end{tabular}}

\end{table}

The Qwen-Image fp16 row uses H100 NVL images at 512$\times$512, reported separately from the 768-pixel selected-pair study. Each of the 192 pairs compares a batch-8 image with the same-prompt solo image from the same stage: 64 pairs each at P10, P25, and P100, weighted equally. Prompts overlap across stages. Metrics use RGB after discarding PNG alpha without compositing. Exact means bit-identical decoded RGB; none of the 192 pairs is exact. SSIM and LPIPS were independently recomputed on CPU.

\textbf{Key observations.} Three observations hold across the table.
\begin{enumerate}
\item The batch-1 rerun is bit-exact while the batched image differs in nearly every case: batching, not the seed, decides the output.
\item The size of the effect depends on the model and the precision. It spans from bit-invariant (HiDream-O1 in bf16 and int8) to structural changes (Janus-Pro-7B, Emu3.5 in bf16 and int8, HiDream-O1 in fp8, Qwen-Image in bf16).
\item In most cases the batched image is a different but equally valid image; the exceptions are discussed in Section IV-C.
\end{enumerate} 

\begin{figure}[!t]
\centering
\includegraphics[width=\columnwidth]{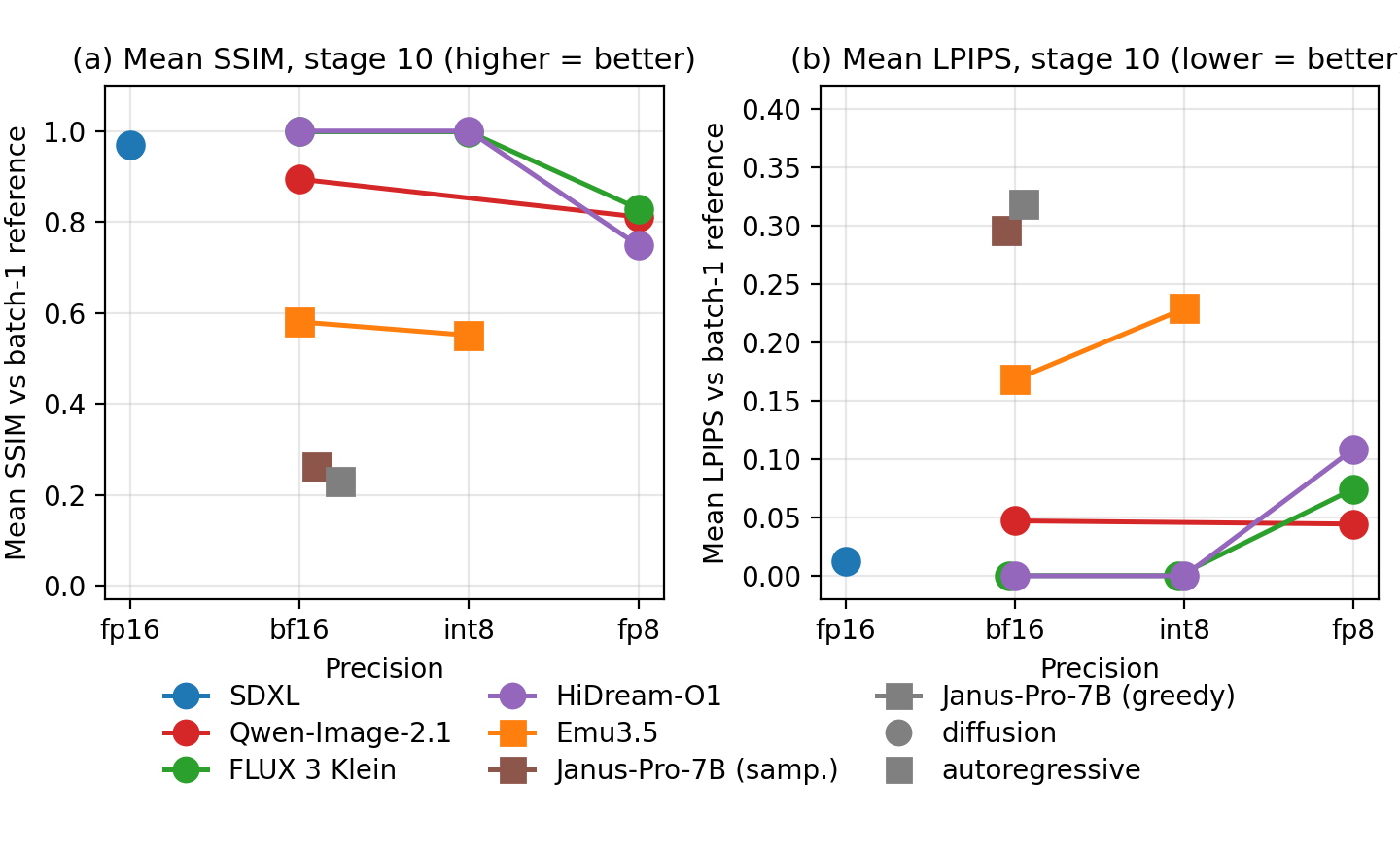}
\caption{Effect of the precision at stage 10, the only stage all six models share: mean SSIM (a) and mean LPIPS (b) of the batched images against the batch-1 reference. Circles: diffusion models; squares: autoregressive models. LPIPS is shown for all six models.}
\label{fig:precision}
\end{figure}

\textbf{SDXL in fp16.} SDXL in fp16 changes every batched image, but the changes stay small. The runner logs a pooled arm-B MAE of 1.85 over 256 pairs at stage 10 (worst pair: prompt 08, MAE 4.98) and 2.164 over 632 pairs at stage 25 (worst pair: prompt 23, MAE 8.72). No arm-B image is bit-identical at either stage. Within arm B, same-prompt pairs diverge in 197 of 3150 pairs at stage 10 and 0 of 7675 pairs at stage 25. The local LPIPS analysis gives arm-B means of 0.013 at both stages ($n = 256$ and $n = 632$), arm-C means of 0.011 and 0.007 against the batch-1 reference ($n = 16$ each), and exactly 0 for the solo-arm repeats ($n = 20$ and $n = 50$): the solo reruns are bit-exact. The worst arm-B pairs reach LPIPS 0.045 (stage 10, prompt 06) and 0.097 (stage 25, prompt 23). Figure~\ref{fig:precision} summarizes the dependence on the precision at stage 10, the only stage that all six models share.

\textbf{SDXL in bf16 at stage 100.} In bf16 at stage 100 (one set of 100 prompts, 313 batch rounds of 8), every one of the 2504 batched images differs from the batch-1 reference (0 of 2504 exact), with mean LPIPS 0.072 and maximum 0.359 (prompt 46). For the same 2504 batch-versus-own-prompt-A0 pairs, mean SSIM is 0.876 (minimum 0.464, maximum 0.982) and mean MAE is 8.854 (minimum 1.755, maximum 42.133). SSIM uses native 1024-pixel RGB arrays, data range 255 and channel averaging; MAE uses all RGB channels on the 0--255 scale. The pooled means weight each pair equally: 96 prompts have 25 pairs and four prompts have 26.

\textbf{Qwen-Image on RTX 6000 Ada.} A separate bf16 stage-100 run at 512$\times$512, 40 steps, has 272 images and no failed-image flags. All 100 solo repeats match their solo references; all 64 batched images and all 8 replay images differ from solo, while all 8 replay images match the corresponding round-0 batch. There are no within-batch repeat pairs at this stage, so companion invariance is not tested by this run. The run takes 2236.5 s, with peak reserved memory of 48,932,847,616 bytes and sampled own-process memory of 47,180 MiB. The distinct GPU/runtime means these counts are reported separately from the H100 metric rows. CPU scoring of the 64 batch-versus-solo pairs gives pooled SSIM 0.894 and LPIPS 0.070 on RGB after discarding alpha without compositing. These metrics belong to this historical 512-pixel run, not the separate fixed-input 768-pixel comparison.

\textbf{Qwen-Image at 512 pixels.} Qwen-Image at 512$\times$512 shows a clear batch effect in both precisions, and the replay arm confirms it is deterministic: arm B holds $n = 64$ images at every stage, arm C replays the round-0 batch with $n = 8$ bit-exact images, and no arm-B image is bit-identical at any stage. SSIM is computed on RGB after discarding the alpha channel. In bf16/fp8 order, the arm-B means are 0.894/0.811 at stage 10 (per-prompt 0.898/0.817, minima 0.667/0.135), 0.850/0.820 at stage 25 (per-prompt 0.850/0.828, minima 0.104/0.131), and 0.875/0.863 at stage 100 (minima 0.268/0.520). The arm-C means are 0.901/0.826 at stage 10 (minima 0.667/0.298), 0.877/0.882 at stage 25 (minima 0.661/0.810), and 0.924/0.951 at stage 100 (minima 0.716/0.920).

\textbf{FLUX 3 Klein.} FLUX 3 Klein, with arms A and B only, stays near-identical in bf16 and int8: mean SSIM about 0.998 in both at stage 10 (B $n = 64$) and about 0.999 in both at stage 25 ($n = 96$), with minima about 0.989 in both at both stages. Only fp8 breaks the pattern: the mean SSIM drops to 0.829 at stage 10 (per-prompt 0.836, minimum 0.260) and 0.855 at stage 25 (per-prompt 0.865, minimum 0.025), with no image bit-identical at either stage.

\textbf{HiDream-O1.} HiDream-O1 is the only model that stays bit-invariant under batching: in bf16 and int8, every batched image equals its batch-1 reference at every stage (all 128 set-3 images bit-identical, SSIM 1.000; all 161 scored LPIPS pairs exact at stage 25; arms A ($n = 100$), B ($n = 128$), and C ($n = 8$) all bit-identical at stage 100, with SSIM 1.000, MAE 0, and LPIPS 0). Only fp8 differs, and the difference is large and stable: it reproduces identically in three runs. The mean SSIM is 0.725 on set 3 (minimum 0.302), every image different, the worst prompts reaching 0.68 and 0.69. By stage: 0.748 at stage 10 (minimum 0.554), MAE 15.45, LPIPS 0.109 (maximum 0.320); 0.731 at stage 25 (B $n = 128$, minimum 0.296), MAE 17.85, LPIPS 0.159 (maximum 0.652), with arm C ($n = 8$) at 0.731 (minimum 0.646), MAE 15.95, LPIPS 0.128 (maximum 0.259) and prompt 19 worst for both metrics; 0.686 at stage 100 (minimum 0.255), MAE 19.56, LPIPS 0.176 (maximum 0.636), with arm C at 0.697 (minimum 0.318), MAE 24.40, LPIPS 0.235 (maximum 0.590) and prompt 41 again worst for both.

\textbf{Autoregressive models.} The autoregressive models show the largest aggregate effects. Emu3.5 reaches mean SSIM 0.580 in bf16 and 0.551 in int8, and all 10 prompts differ from the batch-1 reference (minima 0.216 with MAE 40.07 and 0.308; mean LPIPS 0.168 and 0.229, maxima 0.496 and 0.472). Janus-Pro-7B on the RTX 6000 Ada (bf16) reaches mean SSIM 0.261 with sampled decoding ($T = 1.0$, cfg 5.0) and 0.229 with greedy decoding (minima 0.127 and 0.080; LPIPS 0.296 and 0.319, maxima 0.474 and 0.595). The replay arm reproduces the batched images exactly, so even with sampling the effect is deterministic, not noise. The effect persists at 100 prompts. The batch-versus-solo LPIPS means are 0.368 sampled and 0.367 greedy on the RTX 6000 Ada, and 0.381 and 0.347 on the RTX A6000 (maxima 0.645 to 0.852). Zero of 100 pairs is exact in each of the four combinations, and every solo repeat and batch replay is exact. Mean SSIM is 0.400 sampled and 0.418 greedy on the RTX 6000 Ada (minima -0.010 and 0.050; mean MAE 42.704 and 42.892, maxima 141.809 and 104.883) and 0.409 sampled and 0.435 greedy on the RTX A6000 (minima 0.069 and 0.067; mean MAE 41.018 and 40.037, maxima 100.840 and 101.873).

\subsection{Worst cases}

\begin{figure}[!t]
\centering
\includegraphics[width=0.21\columnwidth]{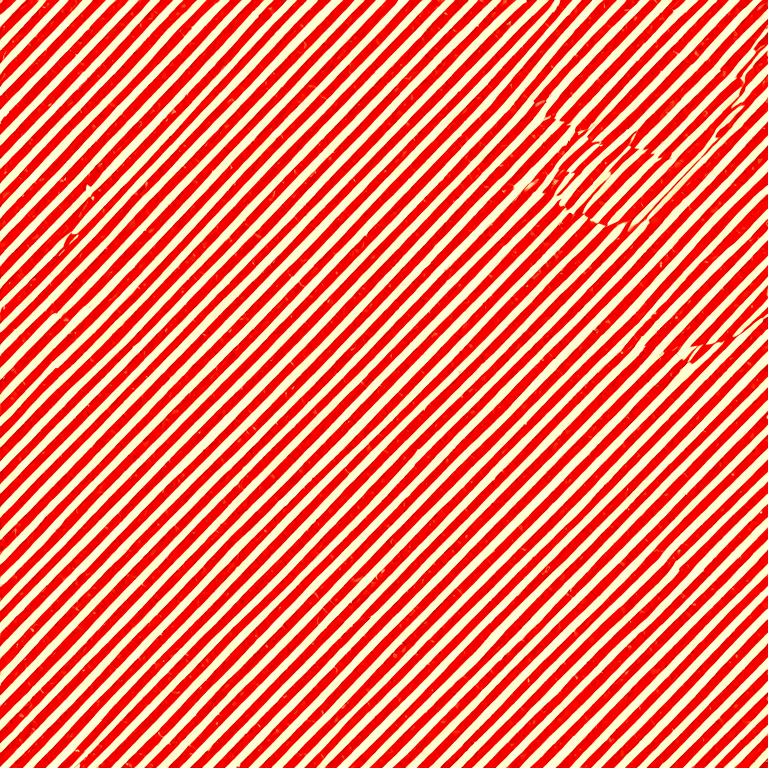}\hspace{0.04\columnwidth}\includegraphics[width=0.21\columnwidth]{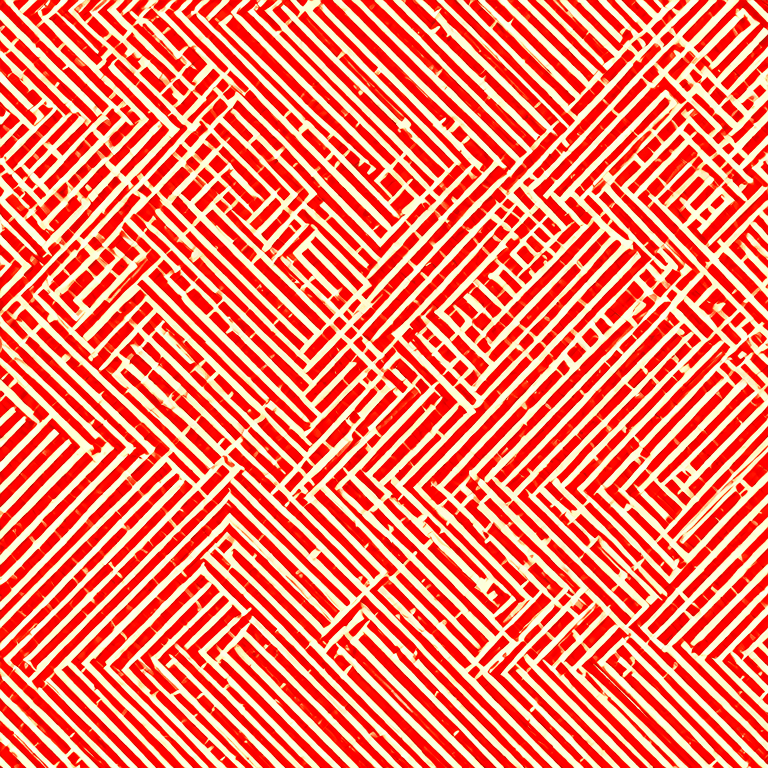}
\caption{Qwen-Image-2.1, bf16, 768 px, H100 NVL, batch 8 versus batch 1. Worst pair (prompt 11, 75-degree red-cream stripes): SSIM 0.228, LPIPS 0.388. Left: batch-1 reference; right: batched image. The same pair is worst on SSIM and LPIPS.}
\label{fig:worst:qwen}
\end{figure}

\textbf{Largest diffusion effect.} The largest diffusion effect we measured is Qwen-Image in bf16 at 768 pixels: for a prompt with 75-degree red-cream stripes, batch size 8 turns the stripes into a chevron crosshatch (SSIM 0.228, LPIPS 0.388), while the same prompt in fp16 stays clean with SSIM 0.997 (Fig.~\ref{fig:worst:qwen}).

\begin{figure}[!t]
\centering
\includegraphics[width=0.21\columnwidth]{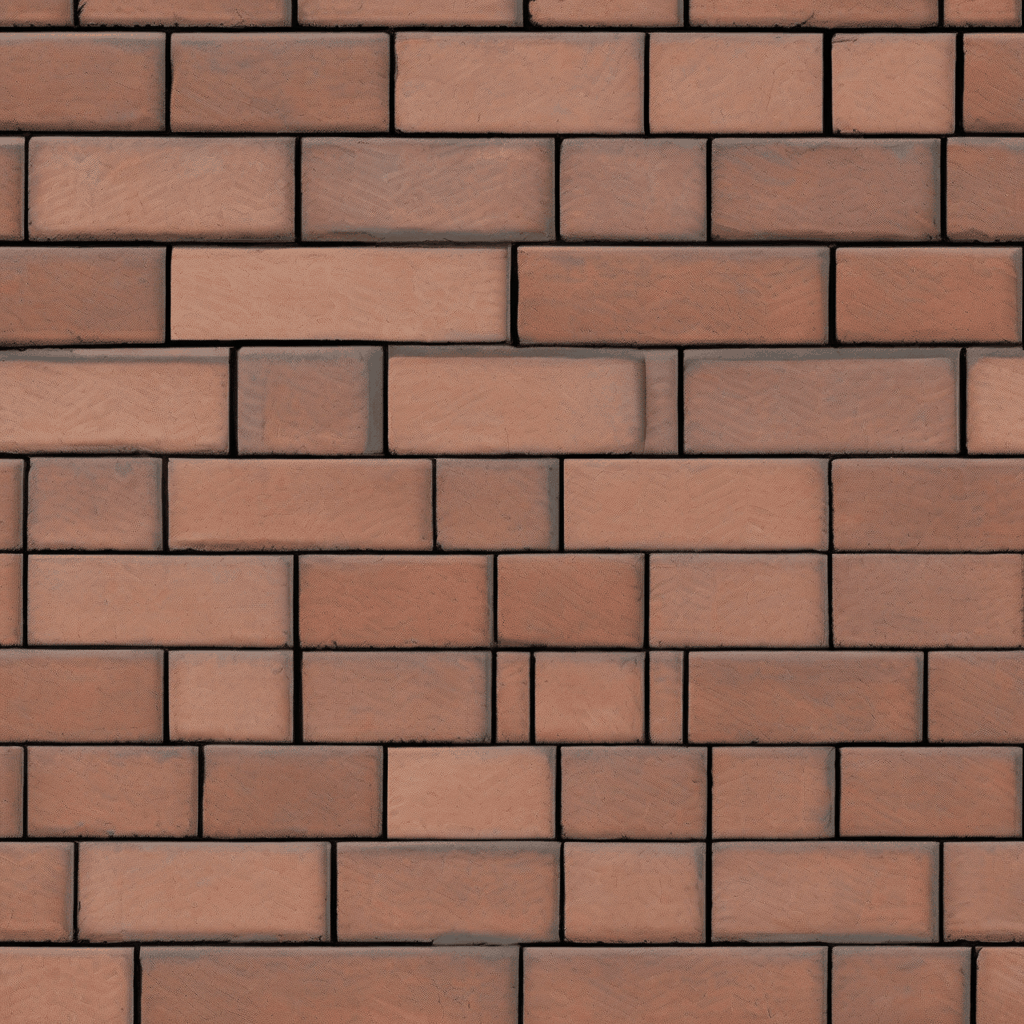}\hspace{0.04\columnwidth}\includegraphics[width=0.21\columnwidth]{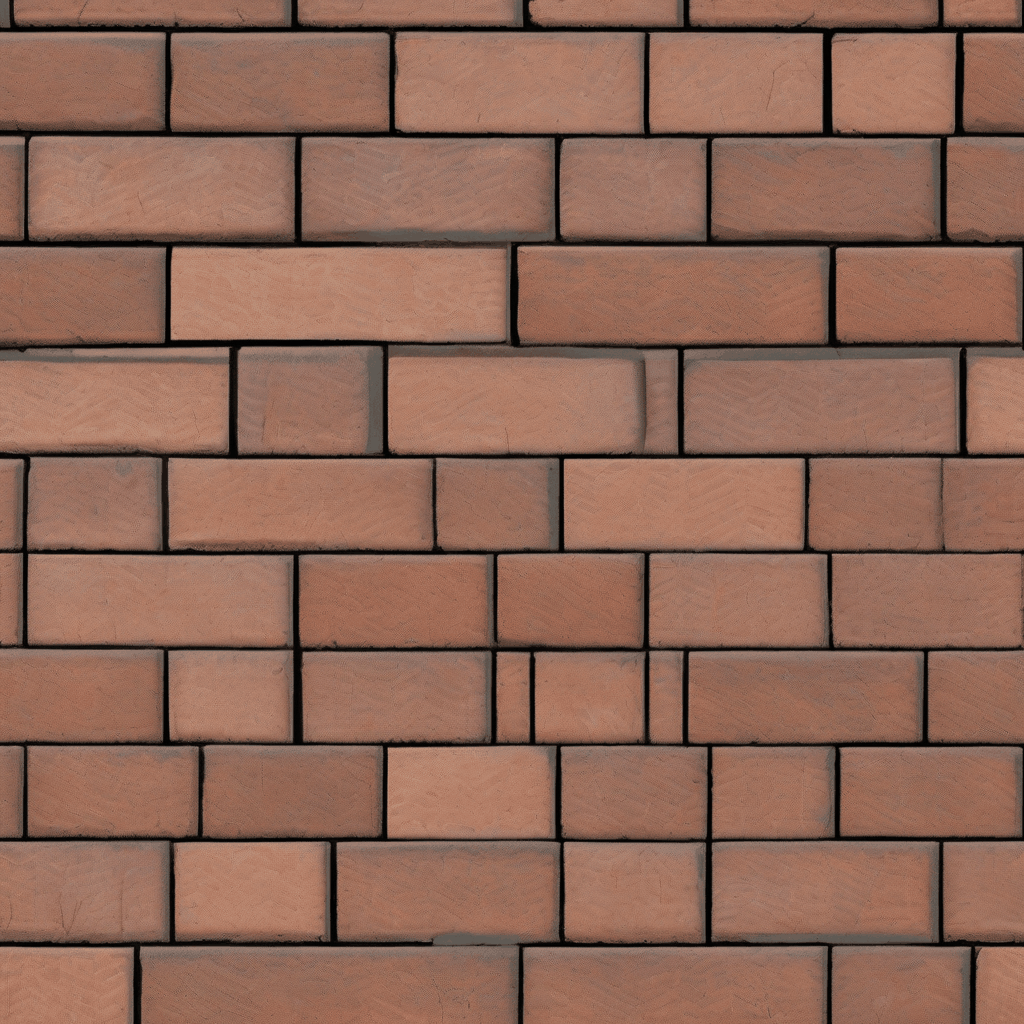}\hspace{0.04\columnwidth}\includegraphics[width=0.21\columnwidth]{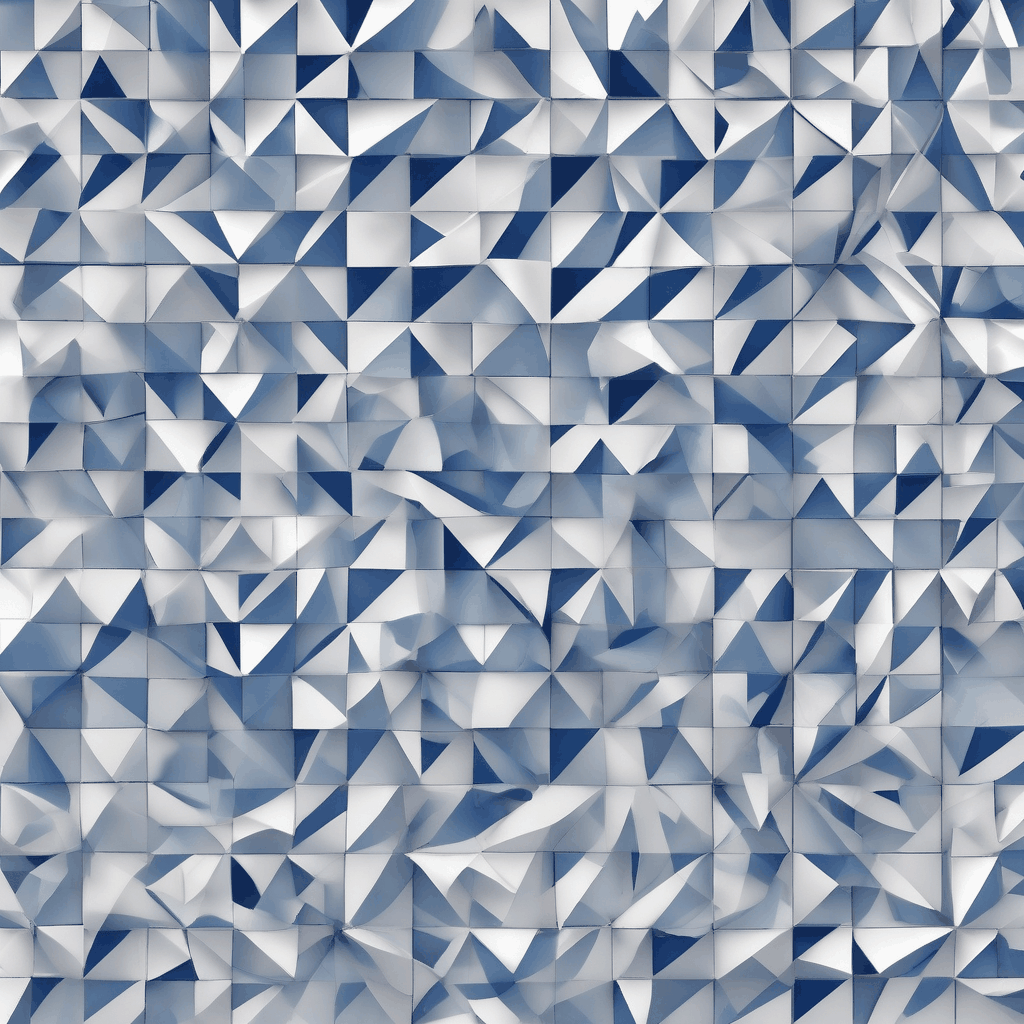}\hspace{0.04\columnwidth}\includegraphics[width=0.21\columnwidth]{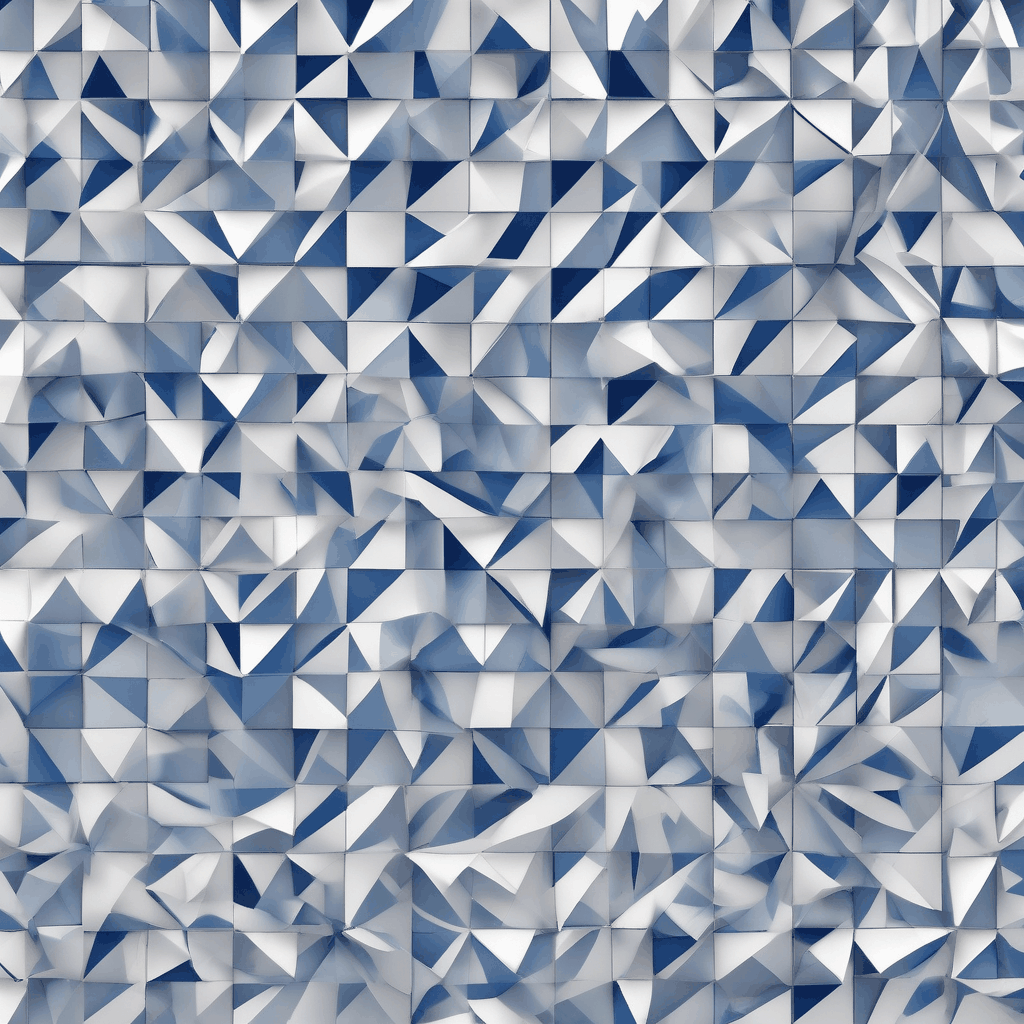}
\caption{SDXL, fp16, RTX A6000, batch versus batch 1. Worst SSIM pair (stage 10, prompt 07, brick wall): SSIM 0.877; worst LPIPS pair (stage 25, prompt 23): LPIPS 0.097. For each pair, left: batch-1 reference; right: batched image.}
\label{fig:worst:sdxl}
\end{figure}

\begin{figure}[!t]
\centering
\includegraphics[width=0.21\columnwidth]{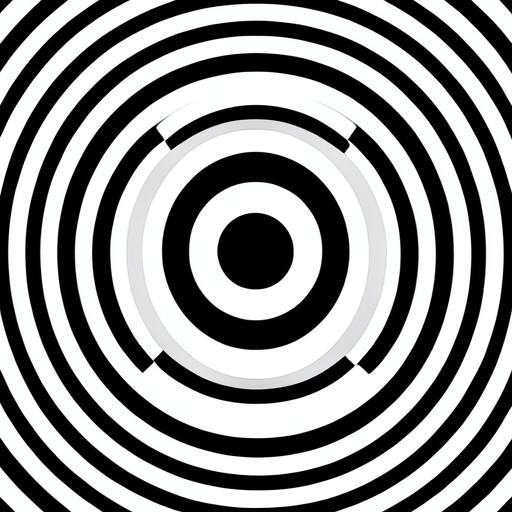}\hspace{0.04\columnwidth}\includegraphics[width=0.21\columnwidth]{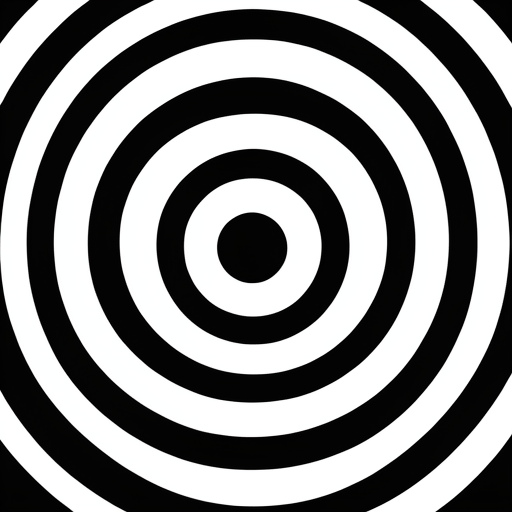}\hspace{0.04\columnwidth}\includegraphics[width=0.21\columnwidth]{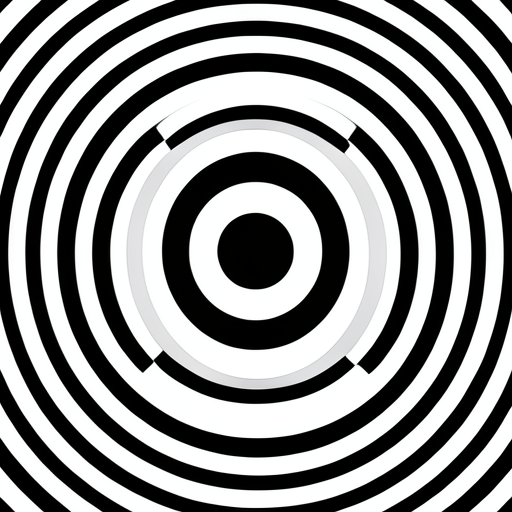}\hspace{0.04\columnwidth}\includegraphics[width=0.21\columnwidth]{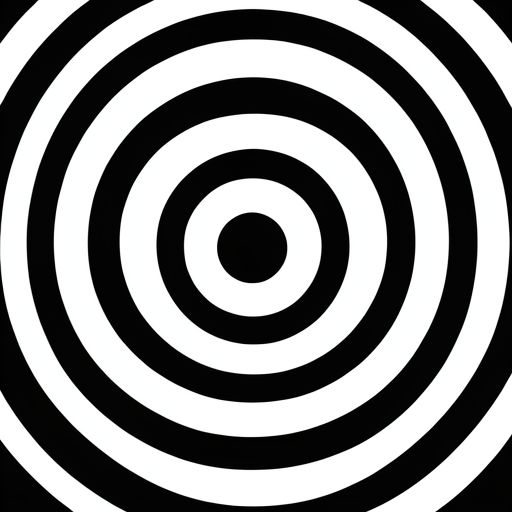}
\caption{FLUX 3 Klein, fp8, H100 NVL, batch versus batch 1. Worst SSIM pair (prompt 00, concentric rings): SSIM 0.260, LPIPS 0.343, MAE 102.5; worst LPIPS pair (prompt 00): LPIPS 0.343.}
\label{fig:worst:flux}
\end{figure}

\begin{figure}[!t]
\centering
\includegraphics[width=0.21\columnwidth]{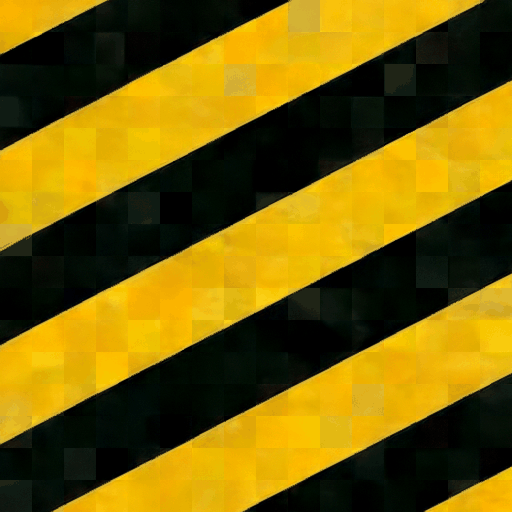}\hspace{0.04\columnwidth}\includegraphics[width=0.21\columnwidth]{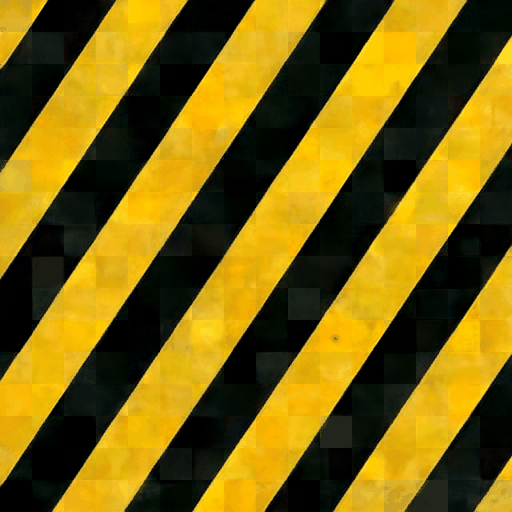}\hspace{0.04\columnwidth}\includegraphics[width=0.21\columnwidth]{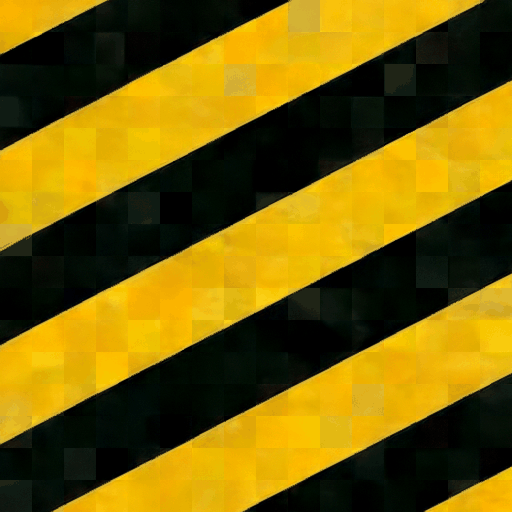}\hspace{0.04\columnwidth}\includegraphics[width=0.21\columnwidth]{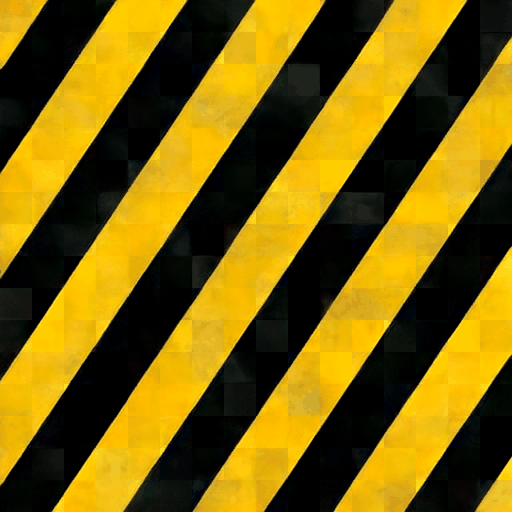}
\caption{HiDream-O1, fp8, H100 NVL, batch versus batch 1. Worst SSIM pair (set 3, prompt 19, angled stripes): SSIM 0.302, MAE 68.97; displayed worst-LPIPS pair (set 3, prompt 19): LPIPS 0.652. The selected SSIM pair's pair-specific LPIPS is 0.644; the 0.652 pair is the archived LPIPS maximum.}
\label{fig:worst:hidream}
\end{figure}

\begin{figure}[!t]
\centering
\includegraphics[width=0.21\columnwidth]{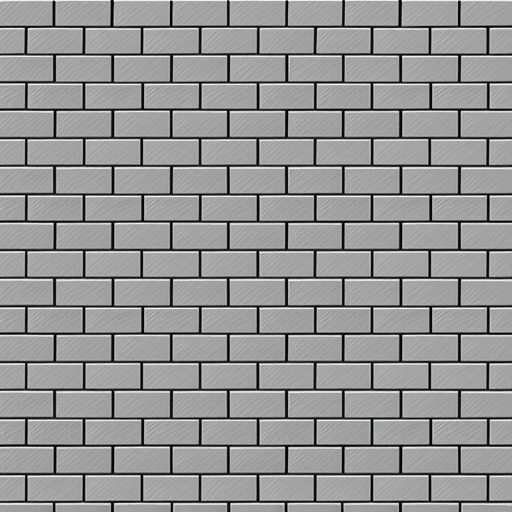}\hspace{0.04\columnwidth}\includegraphics[width=0.21\columnwidth]{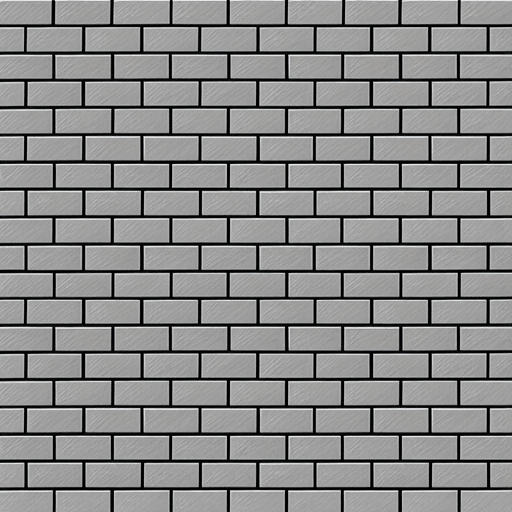}\hspace{0.04\columnwidth}\includegraphics[width=0.21\columnwidth]{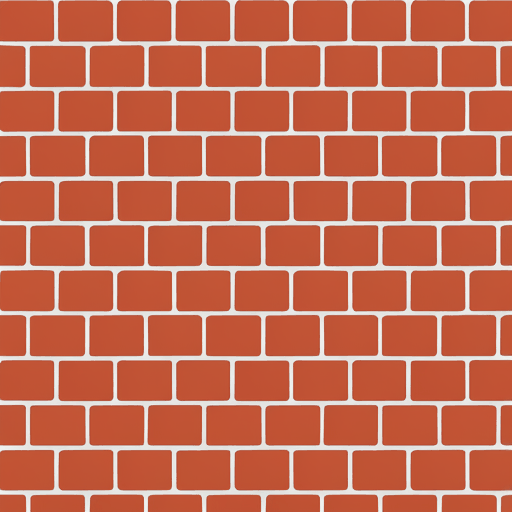}\hspace{0.04\columnwidth}\includegraphics[width=0.21\columnwidth]{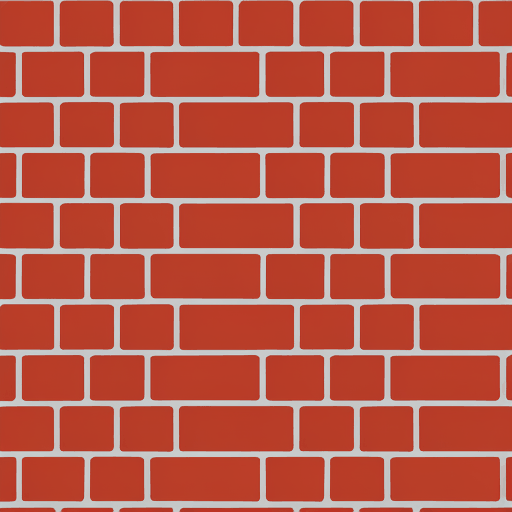}
\caption{Emu3.5, int8, H100 NVL, batch versus batch 1, stage 10. Worst SSIM pair (prompt 5, grey brick wall): SSIM 0.308; worst LPIPS pair (prompt 4): LPIPS 0.472.}
\label{fig:worst:emu}
\end{figure}

\begin{figure}[!t]
\centering
\includegraphics[width=0.21\columnwidth]{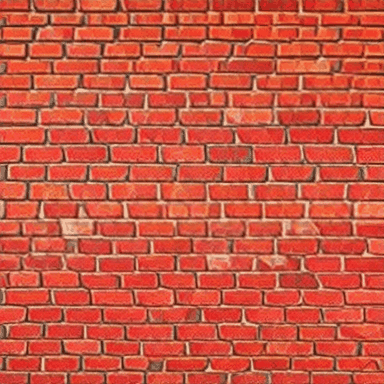}\hspace{0.04\columnwidth}\includegraphics[width=0.21\columnwidth]{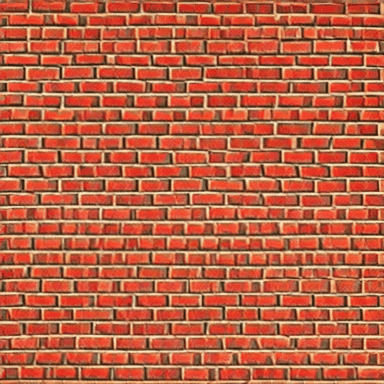}\hspace{0.04\columnwidth}\includegraphics[width=0.21\columnwidth]{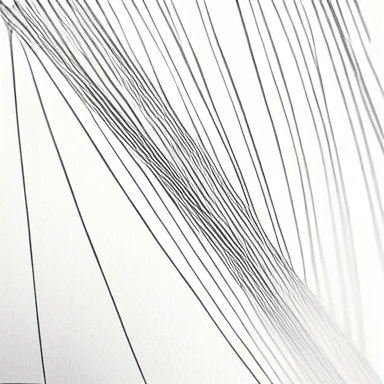}\hspace{0.04\columnwidth}\includegraphics[width=0.21\columnwidth]{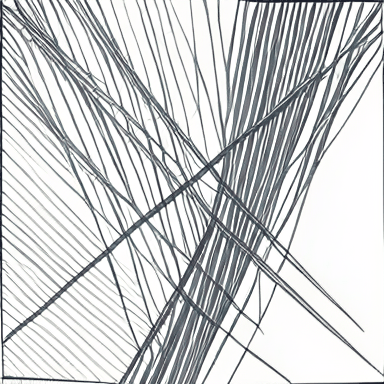}
\caption{Janus-Pro-7B, bf16, RTX 6000 Ada, batch versus batch 1, greedy decoding. Worst SSIM pair (prompt 4, brick wall): SSIM 0.080, LPIPS 0.390; worst LPIPS pair (prompt 8, line pattern): LPIPS 0.595, SSIM 0.122.}
\label{fig:worst:janus}
\end{figure}

\textbf{Digits and text.} Table~\ref{tab:worst} collects the worst cases measured. Figures~\ref{fig:worst:sdxl} to \ref{fig:worst:janus} show, per model, the reference and batched images of the worst-SSIM and worst-LPIPS pairs for the configuration with the largest differences. In most worst cases the batched image is still a plausible image. The exceptions are prompts with digits, text, and counters, where it is visibly worse: a scoreboard rendered by Qwen-Image in bf16 shows wrong or corrupted digits in the batched image while the batch-1 reference is correct (Fig.~\ref{fig:worst:qwen}). For the retained Emu3.5 bf16 seven-segment pair (prompt 102), neither image correctly depicts the requested 88:88. The pair changes with batching, but it is not evidence of a correct solo image becoming incorrect.

\begin{table}[!t]
\centering
\caption{Worst cases measured. Shading as in Table~\ref{tab:main}.}
\label{tab:worst}
\footnotesize\setlength{\tabcolsep}{3pt}
\resizebox{\columnwidth}{!}{%
\begin{tabular}{lllcc}
\hline
Case & Model & Prec. & SSIM & LPIPS \\
\hline
Crosshatch, 768 px, B8 & Qwen-Image-2.1 & bf16 & \cb{0.228} & \cw{0.388} \\
Same prompt & Qwen-Image-2.1 & fp16 & \cg{0.997} & \cg{$\approx$0} \\
Selected SSIM pair, set 3 (p19) & HiDream-O1 & fp8 & \cb{0.302} & \cb{0.644} \\
Earlier sets (means) & HiDream-O1 & fp8 & \cw{\shortstack{0.679\\0.687}} & \cm{\shortstack{0.180\\0.169}} \\
Set 1, p19, r01, slot 0 & HiDream-O1 & fp8 & \cb{0.308} & \cb{0.535} \\
Set 2, p10, r13, slot 6 & HiDream-O1 & fp8 & \cb{0.294} & \cb{0.485} \\
\car{Worst prompt (p5)} & \car{Emu3.5} & \car{int8} & \cb{0.308} & \cw{0.289} \\
\car{Worst prompt (p5)} & \car{Emu3.5} & \car{bf16} & \cb{0.216} & \cw{0.292} \\
Worst prompt, s10 & HiDream-O1 & fp8 & \cw{0.554} & \cw{0.320} \\
Worst prompt, s10 & SDXL & fp16 & \co{0.877} & \cm{0.045} \\
Worst prompt, s10 & FLUX 3 Klein & fp8 & \cb{0.260} & \cw{0.343} \\
Worst prompt, s10 & Qwen-Image-2.1 & bf16 & \cw{0.667} & \cw{0.289} \\
Worst prompt, s10 & Qwen-Image-2.1 & fp8 & \cb{0.135} & \cm{0.083} \\
Worst prompt, s25 (p23) & SDXL & fp16 & \co{0.875} & \cm{0.097} \\
Worst prompt, s25 & FLUX 3 Klein & fp8 & \cb{0.025} & \cb{0.584} \\
Worst prompt, s25 & Qwen-Image-2.1 & bf16 & \cb{0.104} & \cm{0.122} \\
Worst prompt, s25 & Qwen-Image-2.1 & fp8 & \cb{0.131} & \cm{0.083} \\
Worst prompt, s25 (p19) & HiDream-O1 & fp8 & \cb{0.296} & \cb{0.652} \\
Worst prompt, s100 & FLUX 3 Klein & fp8 & \cb{0.274} & \cw{0.343} \\
Worst prompt, s100 & FLUX 3 Klein & bf16 & \cg{0.989} & \cg{0.002} \\
Worst prompt, s100 & FLUX 3 Klein & int8 & \cg{0.989} & \cg{0.002} \\
Worst prompt, s100 & Qwen-Image-2.1 & bf16 & \cb{0.268} & \cw{0.233} \\
Worst prompt, s100 & Qwen-Image-2.1 & fp8 & \cw{0.520} & \cw{0.388} \\
Worst prompt, s100 (p41) & HiDream-O1 & fp8 & \cb{0.255} & \cb{0.636} \\
\car{Worst prompt, sampled} & \car{Janus-Pro-7B} & \car{bf16} & \cb{0.127} & \cb{0.474} \\
\car{Worst prompt, greedy} & \car{Janus-Pro-7B} & \car{bf16} & \cb{0.080} & \cb{0.595} \\
Scoreboard digits & Qwen-Image-2.1 & bf16 & \cw{0.499} & \cb{0.626} \\
\car{Seven-segment digits (p102)} & \car{Emu3.5} & \car{bf16} & \cb{0.254} & \cb{0.470} \\
\hline
\end{tabular}}

\end{table}

The earlier-set HiDream-O1 fp8 row reports unweighted means over prompt means (24 prompts and 128 listed pairs per set), not worst pairs; pooled LPIPS is 0.180 and 0.171 instead. The 125 non-worst pairs per set were checked from per-pair rows rather than rescored from pixels. The Qwen-Image fp16 LPIPS value is positive but rounds to 0.000 (less than 0.001). Qwen-Image pair metrics discard alpha without compositing; the scoreboard SSIM changes to 0.514 on white and 0.328 on black. For the original Emu3.5 p102 solo-versus-B8 pair, SSIM and MAE were independently recomputed. CPU scoring of this retained image pair gives LPIPS 0.470. Neither image correctly depicts the requested 88:88.

For HiDream-O1 fp8, the three selected worst-SSIM pairs in sets 1, 2, and 3 were recomputed with SSIM and LPIPS. The set-3 pair is p19, round 09, slot 1. These are pair-specific scores, not LPIPS maxima over a set. The other 127 batched pairs per set (381 in total) retain archived SSIM values and were not rescored in this check. The earlier-set prompt-mean summary is retained separately.

The SDXL bf16 dtype-probe LPIPS cell in Table~\ref{tab:main} uses 16 batch-versus-solo pairs from eight prompts, with each batch call repeated identically; it is not the stage-100 pool. For the fp16 stage-100 B8 entry in Table~\ref{tab:dose}, SSIM is pooled over 2504 pairs and 100 prompts, with zero bit-identical pairs and a range of 0.875--0.997. Only 100 distinct pair scores occur because batch outputs repeat exactly across rounds; prompts 7, 8, 23, and 99 contain all scores below 0.92.

\subsection{Batch size}

\textbf{Dose-response.} The size of the batch matters much less than its presence: for every model with a batch ladder, the drop from batch 1 to batch 8 is much larger than any change from batch 8 to 32. This section varies only the batch size: the model, precision, prompt set, and seeds are fixed, and the scored pool grows with the batch size as noted below. For FLUX 3 Klein in bf16 the effect does not grow from batch 8 to batch 32, with SSIM at or above 0.989 and LPIPS at or below 0.002 in both cases. Qwen-Image is the only model where pooled SSIM keeps falling as the batch size grows, and the fall is gradual: the pooled SSIM falls in fp8 from 0.863 at batch 8 to 0.837 at batch 16 (minimum 0.197), 0.833 at batch 24 (minimum 0.221), and 0.827 at batch 32 (minimum 0.165), and in bf16 from 0.875 at batch 8 to 0.864 at batch 16 (minimum 0.274) and 0.857 at batch 24 (minimum 0.116). The per-prompt minima fall faster, from 0.520 to 0.165 in fp8 and from 0.268 to 0.116 in bf16. The LPIPS series is not monotone and is reported separately. At batch 32, bf16 reaches pooled SSIM 0.853 and LPIPS 0.089 over 256 batch-versus-solo pairs from 100 prompts in the $512\times512$ H100 ladder. HiDream-O1 in bf16 and int8 stays bit-identical to batch 1 at batches 8, 16, 24, and 32; in fp8 it stays near 0.68 at batch 16 (SSIM 0.676, minimum 0.246, LPIPS 0.181), batch 24 (SSIM 0.683, minimum 0.223, LPIPS 0.176), and batch 32 (SSIM 0.678, minimum 0.204, LPIPS 0.186), nearly unchanged from 0.686 at batch 8. For SDXL, batch 16 reaches CUDA out of memory in the VAE convolution under the 46 GB decimal PyTorch allocator cap on the RTX A6000. Separate batch-24 and batch-32 runs each attempt one batched call after 32 solo batch-1 calls. Neither produces images, and both reach a typed CUDA out-of-memory error under the same cap (42.84 GiB allowed and 16.03 GiB physically free; the batch-32 request is 16.00 GiB), so the failures are cap-bound rather than physical saturation. Batch 8 remains the largest successful measured batch on the 48 GB card; the separate runs' largest successful batch is batch 1 (32 of 32). The pooled SSIM is over $n = 64, 128, 192, 256$ batched images at batch 8, 16, 24, 32 for Qwen-Image-2.1 and $n = 256, 384, 512$ at batch 16, 24, 32 for HiDream-O1, since the batch pool grows with the batch size. The Emu3.5 batch-8 value is the 100-prompt set ($n = 100$). Table~\ref{tab:dose} gives the dose-response values. These comparisons pool more images at larger batches and cover only the three models with a ladder, so the trend is measured on a subset of the matrix.

\begin{table}[!t]
\centering
\caption{Batch-size dose-response. Shading as in Table~\ref{tab:main}. FLUX 3 Klein's batched arm pools batches 8, 16, and 32, so there are no separate B16 and B24 measurements; the effect shows no batch trend. Not attempted means no stage result exists. SDXL B16 and separate batch-24 and batch-32 calls OOM under the 46 GB decimal allocator cap.}
\label{tab:dose}
\footnotesize\setlength{\tabcolsep}{3.5pt}
\begin{tabular}{lllllll}
\hline
Model & Prec. & Metric & B8 & B16 & B24 & B32 \\
\hline
FLUX 3 Klein & bf16 & SSIM & \cg{$\geq$0.989} & \multicolumn{2}{c}{\begin{tabular}[b]{@{}c@{}}\scriptsize pooled in\\[-2pt]\scriptsize B8 and B32\end{tabular}} & \cg{$\geq$0.989} \\
FLUX 3 Klein & bf16 & LPIPS & \cg{$\leq$0.002} & \multicolumn{2}{c}{\begin{tabular}[b]{@{}c@{}}\scriptsize pooled in\\[-2pt]\scriptsize B8 and B32\end{tabular}} & \cg{$\leq$0.002} \\
SDXL & fp16 & SSIM & \co{0.977} & \cellcolor{cbad}OOM & \cellcolor{cbad}OOM & \cellcolor{cbad}OOM \\
Qwen-Image-2.1 & bf16 & SSIM & \co{0.875} & \co{0.864} & \co{0.857} & \co{0.853} \\
Qwen-Image-2.1 & fp8 & SSIM & \co{0.863} & \co{0.837} & \co{0.833} & \co{0.827} \\
Qwen-Image-2.1 & bf16 & LPIPS & \cm{0.088} & \cm{0.093} & \cm{0.091} & \cm{0.089} \\
Qwen-Image-2.1 & fp8 & LPIPS & \cm{0.092} & \cm{0.098} & \cm{0.100} & \cm{0.105} \\
HiDream-O1 & bf16 & SSIM & \cg{1.000} & \cg{1.000} & \cg{1.000} & \cg{1.000} \\
HiDream-O1 & bf16 & LPIPS & \cg{0.0} & \cg{0.0} & \cg{0.0} & \cg{0.0} \\
HiDream-O1 & int8 & SSIM & \cg{1.000} & \cg{1.000} & \cg{1.000} & \cg{1.000} \\
HiDream-O1 & int8 & LPIPS & \cg{0.0} & \cg{0.0} & \cg{0.0} & \cg{0.0} \\
HiDream-O1 & fp8 & SSIM & \cw{0.686} & \cw{0.676} & \cw{0.683} & \cw{0.678} \\
HiDream-O1 & fp8 & LPIPS & \cm{0.176} & \cw{0.181} & \cm{0.176} & \cw{0.186} \\
\hline
\end{tabular}
\end{table}

\subsection{Factor attribution}

\textbf{Batch.} In the diffusion-model experiments, the fixed-batch outputs are stable across the tested companions and positions. A separate Janus-Pro-7B bf16 experiment on the RTX 6000 Ada tests four targets, two overlapping seven-companion sets and target positions 0, 3 and 7, with padded length 52 fixed across solo and batched calls. For both greedy decoding and one fixed sampled seed, all six B8 target outputs per target are identical in tokens and decoded pixels: 120/120 unordered within-target comparisons are exact. All 48 target B8 outputs differ from their solo reference, while all 392 image-level replay pairs are exact. These results support companion and position stability within the tested scope, not across all autoregressive models, prompts or seeds.

\textbf{Model family.} The architecture family separates the effect sizes we measured more than any other factor we varied. The two autoregressive models show the largest effects we measured: Emu3.5 reaches mean SSIM 0.580 in bf16 and 0.551 in int8, and Janus-Pro-7B reaches 0.261 with sampled decoding and 0.229 with greedy decoding, with MAE up to 54.77, mean LPIPS up to 0.319, and every one of the 10 prompts differing from the batch-1 reference in all four autoregressive configurations; at 100 prompts, 0 of 100 batch-versus-solo pairs is exact for both autoregressive models. No diffusion configuration comes close: the smallest diffusion mean SSIM is 0.686 (HiDream-O1, fp8, stage 100), above even the largest autoregressive mean, and the bf16 diffusion runs stay at 0.875 or higher, with HiDream-O1 bit-invariant and FLUX 3 Klein at or above 0.989. The separation is wide: the worst autoregressive mean (0.229) sits about a factor of three below the worst diffusion mean (0.686). The effect is deterministic in both families, and within each family the precision and the model decide the exact value, but the family sets the scale. Section~V discusses a possible mechanism for the difference.

\textbf{Precision.} Reduced precision amplifies the effect, but the bit width alone does not decide it: the worst format is model dependent. The visible cases are specific to 8-bit formats: fp8 in FLUX 3 Klein, HiDream-O1, and Qwen-Image, and int8 in Emu3.5. The worst format is int8 for Emu3.5 (mean SSIM 0.551, against 0.580 in bf16), fp8 for FLUX 3 Klein and HiDream-O1 (whose int8 is bit-invariant on set 3, SSIM 1.000), and bf16 for Qwen-Image-2.1, which is more batch-sensitive than fp16 in 7 of 8 prompts, with fp8 below both. For SDXL, bf16 and int8 both amplify the effect relative to fp16, while bf16 stays close to the reference (SSIM 0.96-0.98) and changes every image. A FLUX 3 Klein int8 probe on a generic RTX A6000 (torchao Int8WeightOnly on the transformer Linear layers, text encoder and VAE in bf16; 50 steps, 28 calls, 148 images, 34.3 minutes, peak 25.916 GiB) reproduces its solo reruns and batch replays exactly (0/10 and 0/64 pairs), but all 64 batch-versus-solo pairs and 132 of 180 within-batch pairs diverge; exactness here means equality of decoded uint8 RGB arrays, not merely equality of PNG file bytes. Mean LPIPS is 0.008 over the 64 batch-versus-solo pairs (maximum 0.051), or 0.007 when each of the 30 distinct decoded pairs is counted once. The 180 unordered same-prompt within-batch pairs include duplicate outputs and comparisons across batch sizes 8, 16, and 32; 52 decoded pairs are distinct. Their mean LPIPS is 0.008 over all listed pairs and 0.010 over the 30 distinct divergent pairs (maximum 0.051). Solo repeats and batch replays have LPIPS 0.000. These pooled results are reported separately from the H100-M rows.

\textbf{GPU.} Changing the GPU changes every matched Janus-Pro-7B bf16 image (RTX A6000 versus RTX 6000 Ada). Over 100 prompts, cross-GPU LPIPS is 0.383 for sampled solo images and 0.368 for batch/replay; greedy values are 0.355 and 0.363. Within-GPU batching gives LPIPS 0.368/0.367 on Ada and 0.381/0.347 on A6000 (sampled/greedy): the effects are of the same order. Table~\ref{tab:factors} summarizes the scope.

In a separate Qwen-Image-2.1 bf16 fixed-input comparison of RTX 6000 Ada and RTX A6000, the cross-GPU difference is smaller than batching over 100 prompts. Across 272 same-arm RGB rows (200 solo, 64 batch and eight replay), mean MAE/SSIM/LPIPS is 2.313/0.970/0.019. These are rows over 100 prompts, not 272 independent samples: 100 solo rows repeat the other solo rows. The 95\% prompt-cluster bootstrap intervals are [1.804, 2.857] for MAE, [0.962, 0.978] for SSIM and [0.012, 0.026] for LPIPS. Restricting the cross-GPU rows to the 64 prompts covered by batching gives 200 rows, SSIM 0.972 and LPIPS 0.019. Batching against each GPU's own solo reference gives SSIM/LPIPS 0.898/0.069 on Ada and 0.907/0.062 on A6000 (64 rows each). Matched per-prompt LPIPS differences, cross-GPU minus batching, are -0.049 [-0.064, -0.036] and -0.042 [-0.057, -0.028], respectively. These deltas average within each prompt before subtraction; they are not differences of pooled means with unequal arm mixtures and row weights. All 100 prompts, including prompt 38, are retained.

This fixed-input cross-GPU difference tests the denoiser and decoder with fixed CPU latents and Ada-produced conditioning, not the text encoder. Each GPU has one run at $768\times768$, 40 steps, bf16 and seed 20261110, with per-image VAE slicing. Runtime, cuDNN and kernel-selection differences are not excluded. cuDNN deterministic mode is off, convolution TF32 is allowed and matrix-multiplication TF32 is off; exact replays do not imply deterministic mode. PNG/native RGBA integrity and RGB metrics are distinct: scoring discards alpha without compositing. Solo reruns are exact for 100/100 prompts and batch replays for 8/8 pairs per GPU; there are no repeated B-versus-B pairs in this stage. No cross-GPU row is exact, an observed exact rate of approximately zero; a bootstrap interval [0,0] is not a population upper bound. These results do not establish an image-quality ranking.

A separate Qwen fp8 full-H100-versus-MIG comparison changes software and runtime too, so it is cross-setting, not GPU-only. None of 92 RGB pairs is exact; solo means are SSIM 0.949, LPIPS 0.018 and MAE 3.888, whereas B8 means round to SSIM 1.000 and LPIPS $\approx$0 (MAE 0.071). Rounded similarity does not imply exact equality.

\begin{table}[!t]
\centering
\caption{Factor attribution per model. Shading as in Table~\ref{tab:main}. The Qwen-Image-2.1 cross-setting comparison is reported in the text.}
\label{tab:factors}
\scriptsize\setlength{\tabcolsep}{2pt}
\resizebox{\columnwidth}{!}{%
\begin{tabular}{llll}
\hline
Model & 16-bit pair & 8-bit formats & Cross-GPU \\
\hline
SDXL & \cm{bf16 amplifies} & \cm{int8 amplifies} & \begin{tabular}[b]{@{}l@{}}fp16 fixed input: 0/302 exact\\MAE 1.455; SSIM 0.987\\LPIPS 0.008\end{tabular} \\
Qwen-Image-2.1 & \cw{bf16 worse, 7/8} & \co{fp8 0.811} & \begin{tabular}[b]{@{}l@{}}bf16, 512 px, p100: SSIM 0.970\\LPIPS 0.019; 0/272 rows exact\end{tabular} \\
FLUX 3 Klein & \cg{bf16 sub-visible} & \co{fp8 0.829} & \begin{tabular}[b]{@{}l@{}}bf16, p100: SSIM 0.699 solo\\0.809 B8; LPIPS 0.139/0.122\end{tabular} \\
HiDream-O1 & \cg{bf16 bit-invariant} & \cg{int8 bit-inv.}; \cw{fp8 0.686} & \begin{tabular}[b]{@{}l@{}}new source, bf16:\\SSIM 0.913 solo\\0.938 B8\end{tabular} \\
\car{Emu3.5} & \begin{tabular}[b]{@{}l@{}}bf16 replay exact;\\fp16 NaN at layer 61;\\no 16-bit factor\end{tabular} & \cw{int8 0.551} \cw{$<$ bf16 0.580} & not measured; no matched run \\
\car{Janus-Pro-7B} & \begin{tabular}[b]{@{}l@{}}AR bf16 vs fp16:\\0/80 token/pixel exact\\fixed bf16 decoder\\SSIM 0.263, LPIPS 0.317\end{tabular} & \begin{tabular}[b]{@{}l@{}}\cb{int8, B8 only: 0/8 bit-id}\\ \cw{LPIPS 0.351 samp.\ / 0.297 greedy}\end{tabular} & \begin{tabular}[b]{@{}l@{}}bf16, p100: 0/100 exact\\ LPIPS 0.355-0.383\end{tabular} \\
\hline
\end{tabular}}

\end{table}

No matched second-GPU Emu3.5 run exists in the study, so cross-GPU attribution is not measured for that model; this absence is not evidence for or against a cross-GPU effect.

The SDXL fixed-input fp16 s10 runs completed 64 generation calls and 302 images on each GPU: 30 A repeats, 32 B rounds and 2 C repeats. The runs use fp16 checkpoint loading and UNet inference with standard force-upcast VAE decode, not uniform all-component fp16. Instrumented wall time is 2027.426 s on the RTX A6000 and 2047.642 s on the RTX 6000 Ada; generation-call time is 1768.860 s and 1788.380 s, and post-processing time is 203.191 s and 203.401 s, respectively. PyTorch peaks are 39.331 GB allocated and 39.456 GB reserved on the A6000, and 39.331 GB allocated and 39.458 GB reserved on the Ada. Sampled nvidia-smi peaks are 39.800 GB (37,956 MiB) own-process and 39.810 GB (37,966 MiB) whole-device on the A6000, and 39.999 GB (38,146 MiB) own-process and 40.009 GB (38,156 MiB) whole-device on the Ada. The sampled peaks are lower bounds. cuDNN deterministic mode was off.

The matched A6000 and Ada outputs differ in every pair. An offline CPU comparison of the 302 same-call, same-tag and same-prompt pairs finds 0 pixel-exact and 0 file-exact matches. Overall mean MAE, SSIM and LPIPS are 1.455, 0.987 and 0.008. Arm means are 1.182/0.991/0.004 for A (30 pairs), 1.485/0.987/0.009 for B (256 pairs), and 1.491/0.988/0.007 for the C replay (16 pairs). The repeated arm outputs make the 302 rows descriptive repeated-output weights, not 302 independent prompt observations.

The Emu3.5 bf16 full batch/replay run completed 24 calls and 40 PNGs. Autoregressive outer wall time is A0 430.868 s, A1 428.476 s, B 128.487 s and R 128.257 s. All 20 replay pairs are exact in tokens, decoded pixels and unnormalized CFG-guided log-scores; these scores are not probability mass, KL or entropy. Maximum PyTorch peaks are 79.399 GB allocated and 87.992 GB reserved, and the sampled nvidia-smi device peak is 87.731 GB (83,667 MiB), a lower bound. This is a bf16 batch/replay result, not a precision comparison; fp16 is unavailable because it produces NaN at layer 61.

Janus compares autoregressive bf16 with fp16 while holding the bf16 image decoder fixed, not an end-to-end fp16 pipeline. The 80 listed contrasts contain 40 unique pairs on 10 prompts, pooled across greedy and sampled decoding; 0 of 80 is bit-exact. SDXL compares batch 8 with batch 1 within each dtype on eight prompts; its 16 rows repeat the same eight pairs. It is not a bf16-versus-fp16 comparison. The SDXL dtypes and the two studies are kept separate. Counts are not independent samples, and no confidence intervals are reported.

In a separate new-source HiDream-O1 bf16 study (not the historical vendor configuration), matched A6000/Ada runs at $512\times512$, 50 steps change every RGB pair. For 100 prompts, cross-GPU solo MAE/SSIM/LPIPS is 5.656/0.913/0.050 (200 listed pairs, 100 unique); B8 gives 4.944/0.938/0.032 (16 listed pairs, eight unique, one composition plus exact replay). The recorded source, weights, stack, settings, initial latents and conditioning match.

For FLUX 3 Klein bf16, a separate matched RTX A6000/RTX 6000 Ada comparison at $512\times512$ and 50 steps gives cross-GPU SSIM/LPIPS 0.699/0.139 for solo images and 0.809/0.122 for B8. The stage-100 solo pool has 200 listed pairs but 100 unique pairs; B8 has 16 listed pairs but eight unique pairs, including replay. No stage-10 results are pooled with these values.

GPU and batch effects are model- and protocol-dependent. The matched Janus results are of the same order; in the fixed-input Qwen p100 comparison, the measured cross-GPU difference is smaller than batching.

\subsection{Root-cause analysis}
Batch shape changes the first observed divergent operation: layer-11 RMSNorm in Janus-Pro-7B bf16, a downsampler convolution in SDXL, and a shared activation-quantization scale in Qwen fp8. Fixed-two-row replay removes the local difference in Janus and SDXL and reduces it in Qwen, but no single intervention restores the complete pipeline. In Janus, combined norm, Linear and attention fixes remove all 576 token mismatches in the tested bf16 trace; the INT8 norm intervention does not generalize to the five tested prompts. Qwen's scale intervention reduces MAE from 2.591 to 1.390 and from 20.240 to 15.880 without restoring solo equality. Batch shape changes specific operations, and fixing one is not enough to make the whole pipeline batch-invariant.

An isolated Janus BF16 RMSNorm replication on the H100 NVL covers 62 positions, 2 through 63, with 8192 elements per position. The captured A6000 batch-1 reference and the H100 batch-1 output match exactly in all 62 positions. Native H100 batch 1 versus batch 8 differs only at position 57, in 11 of 8192 elements, with maximum absolute difference 0.0009765625 and RMS 1.8344e-05. A fixed-two-row batch-8 replay and an fp64 batch-1-versus-batch-8 control are exact in all 62 positions. The fp64 reference differs from the native batch-1 output at position 16 in 18 elements and at position 57 in 11 elements, with the same maximum absolute difference. This result is limited to the isolated RMSNorm replication and does not make a full-model claim.

\subsection{Prompt-type breakdown}

\textbf{Prompt families.} High-frequency patterns near the resolution limit, such as checkerboards, gratings, and angled stripes, include many of the lowest-SSIM cases. Which specific prompt breaks is model dependent, so the per-prompt ranking does not transfer across models. Prompts with digits, text, and counters form a distinct class: there the batched image can be semantically wrong, as in Section IV-C, so the impact is worse, not only different. Table~\ref{tab:prompttypes} summarizes the breakdown.

\begin{table}[!t]
\centering
\caption{Prompt-type breakdown. Shading as in Table~\ref{tab:main}.}
\label{tab:prompttypes}
\footnotesize\setlength{\tabcolsep}{4pt}
\begin{tabular}{ll}
\hline
Prompt family & Finding \\
\hline
High-frequency patterns & \cb{lowest-SSIM cases} \\
Digits, text, counters & \cb{visibly worse, wrong digits} \\
Cross-model transfer & \cm{model-dependent ranking} \\
\hline
\end{tabular}
\end{table}

\section{Discussion}

\textbf{Implications.} The results have direct consequences for how text-to-image models are evaluated and operated. First, a seed does not make a result reproducible. Reporting the prompt and the seed is not enough; the batch size, the precision, and the GPU decide the output as well, so papers and benchmarks should report the serving setup. Second, benchmarks and image-quality comparisons that mix batch sizes, precisions, or hardware can be biased, because part of the measured difference comes from the execution setting rather than the model. Third, audits, A/B tests, and safety evaluations of generated images need the same care: a change in the serving stack can change the images even when nothing about the model changed.

\textbf{Possible causes.} The pattern is consistent with batch-dependent kernel execution: reduction orders change with tensor shapes, and 8-bit formats amplify small numerical differences into visible ones, with fp8 dynamic quantization adding a batch-wide scale that couples the requests in a batch. Confirming this account requires kernel-level instrumentation, which we leave for future work.

\textbf{Why the autoregressive models are more sensitive.} The two autoregressive models show the largest effects we measured (mean SSIM 0.580 and 0.551 for Emu3.5, 0.261 and 0.229 for Janus-Pro-7B), and the replay arm reproduces the batched images exactly for both, so the difference is decided by the batch, not by sampling. A likely explanation is the sequential structure of autoregressive decoding. The image is produced as a long sequence of discrete tokens, each conditioned on all the previous ones, so a small numerical change that flips one token changes the context for every later token, and the perturbation compounds along the sequence. Diffusion models instead update the full latent in parallel at every step, which can average or contract small perturbations instead of propagating them. Two observations fit this account. First, the effect in Janus-Pro-7B is similar for greedy and sampled decoding, which is consistent with flips at near-tied logits rather than with random draws. Second, no diffusion configuration we measured reaches the effect sizes of the autoregressive models. The Janus trace evidence is limited to the tested bf16 trace and prompts, so this account remains a hypothesis.

\textbf{Dependence on quantization.} The bit width alone does not decide the effect. int8 leaves HiDream-O1 bit-invariant at every stage we measured and keeps FLUX 3 Klein sub-visible (SSIM 0.998 at stage 10), while it degrades Emu3.5 below its bf16 value (0.551 versus 0.580). fp8 is the worst precision on every model where we measured it: 0.811 for Qwen-Image, 0.829 for FLUX 3 Klein, and 0.748 for HiDream-O1 at stage 10, dropping to 0.686 at stage 100. A plausible explanation is how each format is executed. The int8 configurations we tested quantize the weights only and dequantize into the bf16 compute path, so the arithmetic kernels and reduction orders match the bf16 run. fp8 changes the compute precision itself, with scaling and rounding executed in kernels whose reduction order depends on the batch shape, and with a batch-wide scale that couples the requests in a batch. Under this account, batching changes little when the compute path is unchanged and more when the batch enters the arithmetic. Emu3.5 does not fit this pattern cleanly, since its int8 configuration is more batch-sensitive than bf16 even though both share the bf16 compute path. Beyond the scoped traces and interventions above, we did not instrument the full kernel set, so these accounts remain hypotheses.

\textbf{Mitigation options.} For settings where reproducibility matters, our measurements suggest three options. First, on the model choice, the diffusion models in our matrix stay well above the autoregressive ones: the lowest diffusion mean SSIM we measured is 0.686 (HiDream-O1, fp8, stage 100), against 0.580 and 0.551 for Emu3.5 and 0.261 and 0.229 for Janus-Pro-7B, though the autoregressive scores use 10-prompt sets against up to 100 for the diffusion models. Second, generate at batch 1. The batch-1 rerun is bit-exact for every model and precision we tested (Table~\ref{tab:replay}), so batch 1 removes the batch effect on a given GPU, at the cost of throughput. It does not make outputs portable: for Janus-Pro-7B, 0 of 10 prompts are bit-identical across the RTX 6000 Ada and the RTX A6000 with the prompt and seed fixed. Third, pick the precision per model, because the ordering does not transfer. fp8 is the worst 8-bit format on every model where we measured it (0.811 for Qwen-Image, 0.829 for FLUX 3 Klein, 0.748 for HiDream-O1 at stage 10) and is the one to avoid. fp16 is the safest choice for SDXL, since bf16 and int8 both amplify the effect, and for Qwen-Image, where bf16 is more batch-sensitive than fp16 in 7 of 8 prompts. FLUX 3 Klein and HiDream-O1 tolerate bf16 and int8 with sub-visible or zero error, and Emu3.5 degrades less in bf16 (0.580) than in int8 (0.551).

\textbf{Limitations.} The study covers six models and three GPU types, with up to 100 prompts per set for most models (SDXL is scored to 25, except bf16, whose LPIPS and exactness reach 100 prompts) and one base seed per prompt set. We measure the effect but do not evaluate a complete serving fix; the local interventions tested do not make the whole pipeline batch-invariant. Batch-invariant execution for image models remains an open problem.

\begin{table}[!t]
\centering
\caption{Impact of the factors studied, with the prompt and seed fixed. Shading as in Table~\ref{tab:main}: green none, yellow small, orange large, red largest.}
\label{tab:impact}
\footnotesize\setlength{\tabcolsep}{3pt}
\begin{tabular}{@{}p{5.6cm}l@{}}
\hline
Factor & Impact \\
\hline
Batch presence (batched versus solo) & \cb{largest} \\
Model family (autoregressive versus diffusion) & \cb{largest} \\
Precision (8-bit formats versus 16-bit) & \cw{large} \\
GPU (RTX 6000 Ada versus RTX A6000, Janus bf16) & \cw{large} \\
Prompt type (high-frequency patterns) & \cw{large} \\
Batch size beyond 8 & \co{small} \\
int8 weight-only versus bf16 (none for HiDream-O1 and FLUX 3 Klein) & \co{small} \\
Exact replay of the same run & \cg{none} \\
Companion prompts and position (tested diffusion and Janus scope) & \cg{none} \\
\hline
\end{tabular}
\end{table}

\section{Conclusion}

Fixing the prompt and the seed does not fix the image on GPUs. Across six diffusion and autoregressive text-to-image models, batching changes nearly every output, and reduced precision often amplifies the change, in some cases to images that are visibly worse. The replay controls reported here are bit-exact, so the effect is a property of the execution setting, not of sampling noise. Batch presence has the broadest effect in our matrix, the model family sets the scale of the effect, and the GPU can change every image; the relative size of the GPU and batch effects is model- and protocol-dependent. Reproducible text-to-image evaluation therefore needs the batch size, the precision, and the GPU reported alongside the prompt and the seed.

\section*{Acknowledgment}
This work was supported by the Agencia Estatal de Investigación (AEI) (doi:10.13039/501100011033) under Grant ANNA (PID2025-169490OB-C21), by TUCAN6-CM (TEC-2024/COM460) funded by CM (ORDEN 5696/2024) and by European Union's Horizon Europe research and innovation programme under project BRIDGE-AI (Grant 101299050).

\bibliographystyle{IEEEtran}
\footnotesize
\bibliography{refs}


\vspace{-25pt}
 
\begin{IEEEbiographynophoto}{Javier Conde} received his Ph.D. in Telematics at Universidad Politécnica de Madrid, Madrid, Spain. Now is an associate professor at the same university. His research interests include Digital Twins, Big Data, Semantic Web, and Artificial Intelligence.
\end{IEEEbiographynophoto}

\vspace{-25pt}

\begin{IEEEbiographynophoto}{Xiuyuan Qi}
received the B.S. degree in electronic information engineering from Harbin Engineering University, Harbin, China, in 2022. He is currently pursuing the Ph.D. degree with University of Electronic Science and Technology of China, Chengdu, China. His current research interests include fault-tolerant design of low-power AI processors and computer architecture design.
\end{IEEEbiographynophoto}

\vspace{-25pt}

\begin{IEEEbiographynophoto}{Shanshan Liu}
received the Ph.D. degree in Microelectronics and Solid-State Electronics from Harbin Institute of Technology, China, in 2018. She was a post-doctoral researcher at Northeastern University, Boston, USA, from 2018 to 2021, and an assistant professor at New Mexico State University, Las Cruces, USA from 2021 to 2023. She is currently a professor with the School of Information and Communication Engineering, University of Electronic Science and Technology of China.
\end{IEEEbiographynophoto}

\vspace{-15pt}

\begin{IEEEbiographynophoto}{Pedro Reviriego}
received the M.Sc. and Ph.D. degrees in Telecommunications engineering from the Technical University of Madrid, Madrid, Spain, in 1994 and 1997, respectively. From 1997 to 2000, he was an R\&D Engineer with Teldat, Madrid. In 2000, he joined Massana and from 2004 to 2007, he was a Distinguished Member of Technical Staff with the LSI Corporation. From 2007 to 2018 he was with Universidad Antonio de Nebrija, and from 2018 to 2022 with Universidad Carlos III de Madrid. He is currently with Universidad Politécnica de Madrid.
\end{IEEEbiographynophoto}

\vspace{\baselineskip}
\vspace{\baselineskip}

\end{document}


\appendices

\section{Model checkpoints and tooling}

Table~\ref{tab:repos} lists the exact checkpoints used, and Table~\ref{tab:tooling} the main software tooling. The FLUX 3 Klein base repository is gated; the license was accepted before download. All links were verified on 2026-11-11.

\setcounter{table}{0}
\renewcommand{\thetable}{A\arabic{table}}

\begin{table*}[!t]
\centering
\caption{Model checkpoints (Hugging Face repositories).}
\label{tab:repos}
\footnotesize\setlength{\tabcolsep}{4pt}
\begin{tabular}{ll}
\hline
Model & Repository \\
\hline
SDXL base 1.0 & \url{https://huggingface.co/stabilityai/stable-diffusion-xl-base-1.0} \\
Qwen-Image-2.1 & \url{https://huggingface.co/Qwen/Qwen-Image-2.1} \\
FLUX 3 Klein base & \url{https://huggingface.co/black-forest-labs/flux-3-image-klein-base} \\
HiDream-O1-Image & \url{https://huggingface.co/HiDream-ai/HiDream-O1-Image} \\
Emu3.5-Image & \url{https://huggingface.co/BAAI/Emu3.5-Image} \\
Janus-Pro-7B & \url{https://huggingface.co/deepseek-ai/Janus-Pro-7B} \\
\hline
\end{tabular}
\end{table*}

\begin{table}[!t]
\centering
\caption{Main software tooling.}
\label{tab:tooling}
\footnotesize\setlength{\tabcolsep}{4pt}
\begin{tabular}{ll}
\hline
Tool & Repository \\
\hline
PyTorch & \url{https://pytorch.org} \\
diffusers & \url{https://github.com/huggingface/diffusers} \\
transformers & \url{https://github.com/huggingface/transformers} \\
torchao & \url{https://github.com/pytorch/ao} \\
scikit-image & \url{https://scikit-image.org} \\
LPIPS & \url{https://github.com/richzhang/PerceptualSimilarity} \\
\hline
\end{tabular}
\end{table}

\section{Experiment details}
\label{app:expdetails}
\textcolor{black}{This appendix describes the common experimental protocol and model-specific settings. Additional visual comparisons are provided in Appendix C.}

\textbf{Common protocol.} For each model, precision, and prompt: arms A0 and A1 generate the image alone at batch size 1; arm B generates it inside a batch with a cell of 8 prompts, mixing prompts in rotating order; for SDXL, Qwen-Image-2.1, and HiDream-O1, arm C repeats the round-0 batch verbatim as the replay; FLUX 3 Klein, Emu3.5, and Janus-Pro-7B use arm R, which replays arm B. The batch-1 reference is A0 (r0). Prompts are organized in $S$ sets of $P$ prompts with one base seed per set and different seeds across sets. Metrics: fraction bit-identical (decoded pixel arrays equal), MAE on a 0 to 255 scale, SSIM (scikit-image 0.26.0, RGB, native resolution, data range 255, channel average), and LPIPS (lpips 0.1.4, AlexNet backbone, inputs scaled to $[-1, 1]$). The full study runs in three nested stages ($P = 10$, 25, and 100 prompts per set, one seed per set); each stage is a checkpoint whose results remain valid subsets of the final run. Run times and memory use are given in Appendix~\ref{app:schedule}.

\textbf{SDXL base 1.0.} Precisions fp16 and bf16 in a single configuration; 1024$\times$1024, 30 steps, guidance 7.5; $S = 3$, $P = 10$, 25, and 100 over the nested stages; arm B = 256 (stage 10), 632 (stage 25), and 2504 (stage 100) images, arm C = 16; arm A runs 3 identical solo repeats (r0, r1, r2) of $P$ images each (30 images at stage 10, 75 at stage 25, and 300 at stage 100); generated PNGs per run: 302 at stage 10, 723 at stage 25, and 2820 at stage 100. GPU: RTX A6000. Software: torch 2.6.0 (CUDA 12.4), diffusers 0.31.0, transformers 4.46.3, numpy 1.26.4. The stage-2 run sets the PyTorch allocator flag expandable\_segments:True, which changes memory allocation only and not numerics. Full run: 1.22 h (M).

\textbf{Qwen-Image-2.1.} Precisions bf16 and fp8; $S = 2$; 768$\times$768 for the selected-pair study; the historical H100 stage series and completed B32 result are at 512$\times$512 and are reported separately; 40 steps (pipeline default), guidance off (true\_cfg\_scale 1.0), base seed 20261110; arm B = 64 at every stage (8 rounds of 8), arm C = 8, and arms A0 and A1 hold $P$ images each; generated PNGs per run: 92 at stage 10, 122 at stage 25, and 272 at stage 100, plus 416 for the stage-100 B24 ladder arm (B = 192, C = 24); the stage-100 B16 ladder arm holds 344 generated PNGs in each precision (A = 200, B = 128, C = 16), the B32 arm holds 488 in fp8 (A = 200, B = 256, C = 32). The completed bf16 B32 result reported here is a separate $512\times512$ H100 run with 256 batch-versus-solo pairs over 100 prompts; it is not a continuation of a 768-pixel recipe. GPU: H100 NVL. Software: torch 2.14.1, diffusers 0.41.0, transformers 5.20.2, torchao 0.18.0.

\textbf{FLUX 3 Klein base.} Precisions bf16, int8 weight-only (torchao), and fp8 (torchao float8); 512$\times$512, 50 steps, guidance 3.5; $S = 3$, $P = 10$, 25, and 100 over the nested stages; arm R replaces arm C and replays arm B; stage sizes, the same in all three precisions: A0 = 10, A1 = 10, B = 64, R = 64 at stage 10 (148 generated PNGs per precision), A0 = 25, A1 = 25, B = 96, R = 96 at stage 25 (242), and A0 = 100, A1 = 100, B = 344, R = 344 at stage 100 (888); batch ladder 8, 16, 32 configured; 2 to 3 identical-config repeats per cell. GPU: H100 NVL (MIG 3g.47gb). Software: torch 2.15.0 (CUDA 13.2), diffusers 0.42.1, transformers 5.22.2, torchao 0.18.0, numpy 2.4.6.

\textbf{HiDream-O1-Image.} Precisions bf16, int8 (torchao Int8WeightOnly on the language-model linears), and fp8; 512$\times$512, 50 steps, guidance 5.0; $S = 3$, $P = 10$, 25, and 100 over the nested stages; arms A0 and A1 hold $P$ images each; at batch 8, B = 128 and C = 8 at every stage (156 generated PNGs at stage 10, 186 at stage 25, 336 at stage 100); the stage-100 batch ladder adds batch 16 (B = 256, C = 16, 472), batch 24 (B = 384, C = 24; 608 generated PNGs, 508 scored pairs against A0), and batch 32 (B = 512, C = 32, 744); 2 to 3 identical-config repeats per cell. GPU: H100 NVL (MIG 3g.47gb). Software: torch 2.14.1 (CUDA 13.0), diffusers 0.39.0, transformers 4.57.1, torchao 0.18.0. The original HiDream-O1 generation environment did not record Python or numpy versions, so they are not reported. The newer matched-source environment is separate and is not substituted for the historical configuration.

\textbf{Emu3.5-Image.} 34B autoregressive model, greedy argmax decoding (no sampling seed); the language-model stack uses bf16 compute, with an int8 variant that uses torchao weight-only quantization on 448 quantized linears; $P = 10$, $S = 1$; arms A0, A1, B (cell of 8), and R hold 10 images each, 40 generated PNGs per run at stage 10, and 100 images each at stage 100 in bf16 and int8, 400 per run; batch size 8, guidance 2.0, 32$\times$32 token grid (512$\times$512 images), seed 20261110, eager attention. GPU: H100 NVL full device (bf16 needs more than 48 GB). Software: torch 2.14.1 (CUDA 12.6), Python 3.10. An int8 repeat reproduces the run exactly.

\textbf{Janus-Pro-7B.} Autoregressive model; 384$\times$384 images, 576 image tokens; $P = 10$ (same prompts as Emu3.5), $S = 1$; greedy (argmax) and sampled ($T = 1.0$, cfg 5.0) decoding; the batched arm uses 2 cells (8+2); the greedy arm names G0/G1/GB/GR are mapped to A0/A1/B/R for scoring; each arm holds 10 images per decoding mode, 80 generated PNGs per run (40 per mode) at stage 10, and 100 images per mode, 800 per run, at 100 prompts. Stage 10 ran on one RTX 6000 Ada in bf16 with sdpa attention; seed 20261110. The cross-GPU arm ran on one RTX A6000; the 100-prompt bf16 runs completed on both GPUs.

\section{Additional configuration comparisons}


\textcolor{black}{Figures A1–A5 compare selected batched outputs with their corresponding batch-size-1 references.}
One placeholder slot per additional configuration that shows relevant differences (layout as in Figures 3 to 8: reference, worst-SSIM image, reference of the worst-LPIPS pair, worst-LPIPS image). Configurations with identical outputs (SSIM $= 1$, LPIPS $= 0$), such as HiDream-O1 in bf16 and int8, get no image slot, only the table row.

\setcounter{figure}{0}
\renewcommand{\thefigure}{A\arabic{figure}}

\begin{figure}[H]
\centering
\includegraphics[width=0.21\columnwidth]{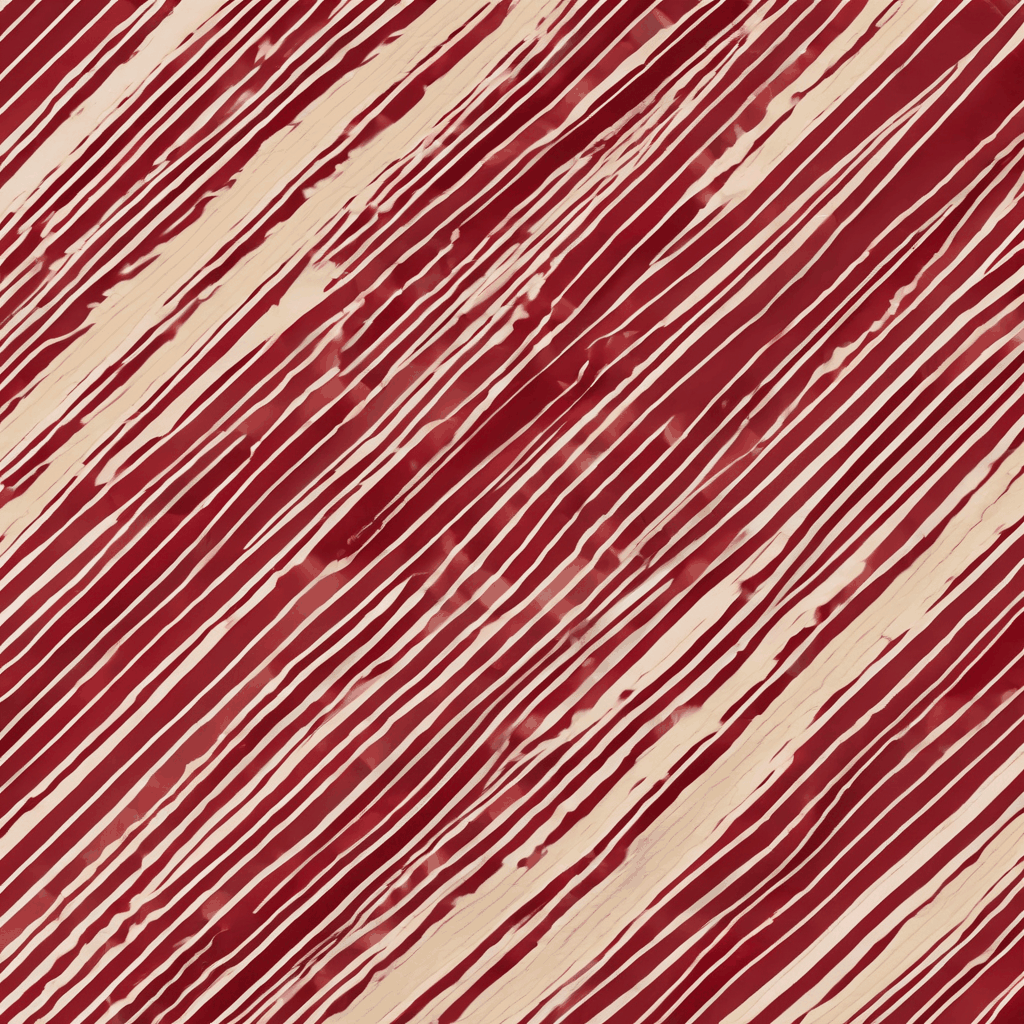}\hspace{0.04\columnwidth}\includegraphics[width=0.21\columnwidth]{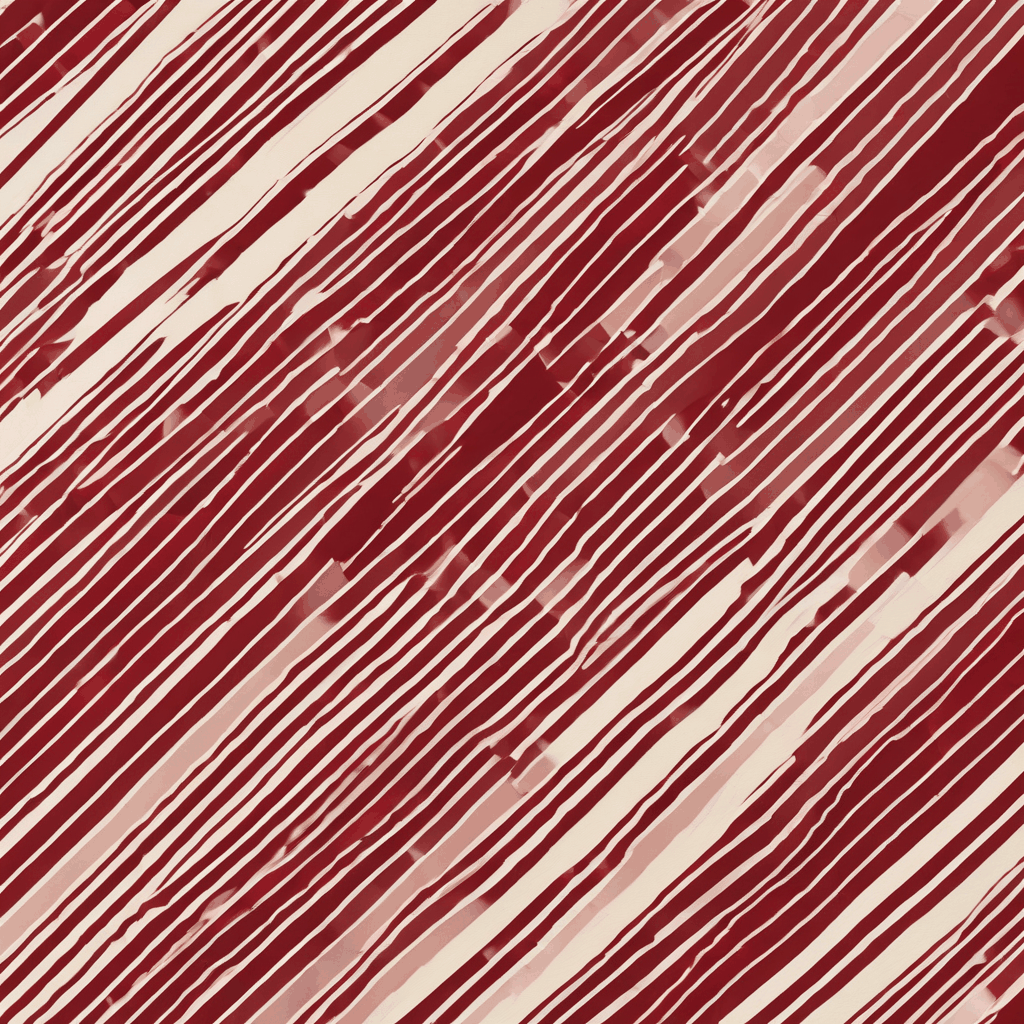}
\caption{SDXL, bf16, RTX A6000, batch 8 versus batch 1 (dtype probe, 8 prompts). Worst pair by mean absolute difference (prompt 02): MAE 24.10 (0-255); SSIM 0.685, LPIPS 0.119.}
\label{fig:annex:sdxlbf16}
\end{figure}

\begin{figure}[H]
\centering
\includegraphics[width=0.21\columnwidth]{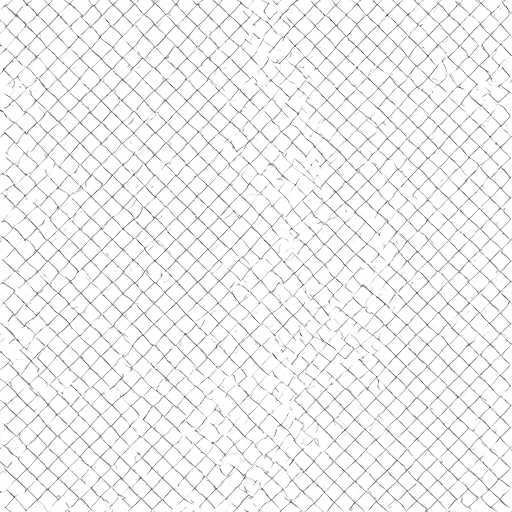}\hspace{0.04\columnwidth}\includegraphics[width=0.21\columnwidth]{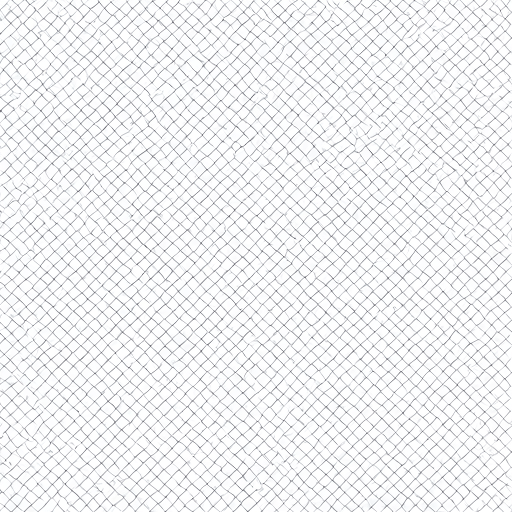}\hspace{0.04\columnwidth}\includegraphics[width=0.21\columnwidth]{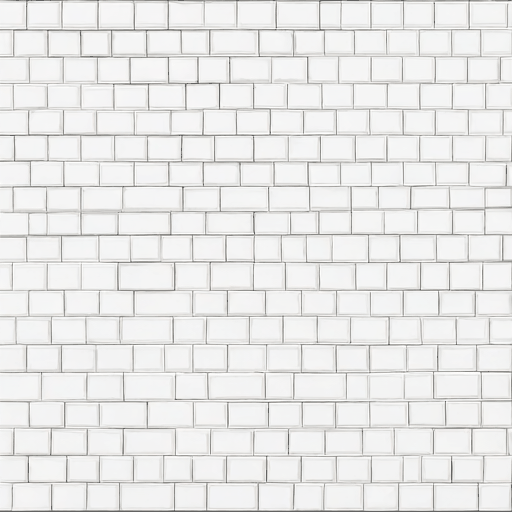}\hspace{0.04\columnwidth}\includegraphics[width=0.21\columnwidth]{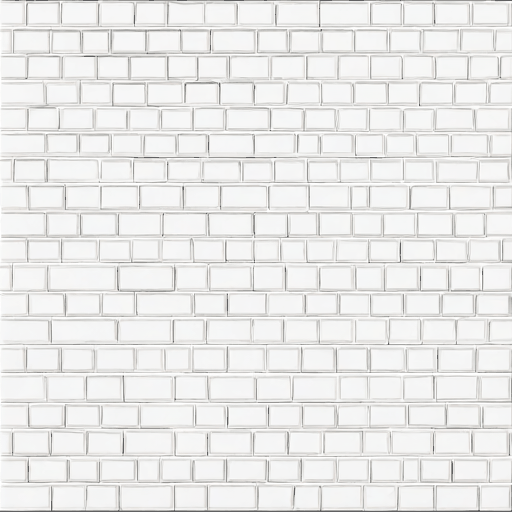}
\caption{Qwen-Image-2.1, fp8, H100 NVL. Worst SSIM pair (prompt 09, fine grid): SSIM 0.135, LPIPS 0.083, MAE 20.9; worst LPIPS pair (prompt 06): LPIPS 0.152, SSIM 0.685, MAE 9.58.}
\label{fig:annex:qwenfp8}
\end{figure}

\begin{figure}[H]
\centering
\includegraphics[width=0.21\columnwidth]{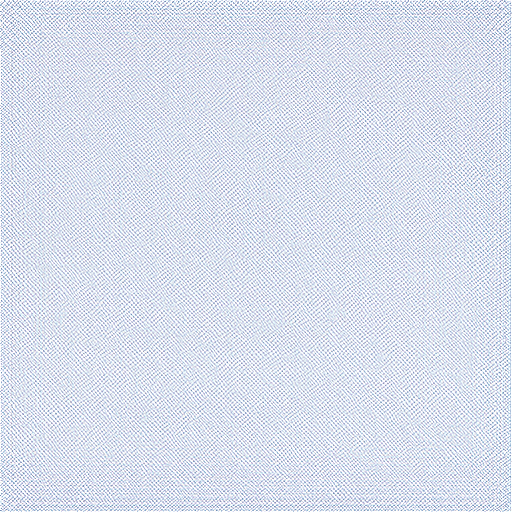}\hspace{0.04\columnwidth}\includegraphics[width=0.21\columnwidth]{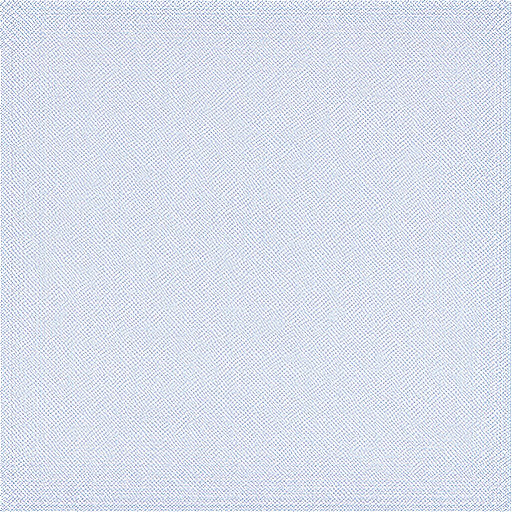}
\caption{FLUX 3 Klein, bf16, H100 NVL. Worst SSIM pair (prompt 09): SSIM 0.989, LPIPS 0.002, MAE 2.37; the worst-LPIPS pair is the same pair (prompt 09, ties).}
\label{fig:annex:fluxbf16}
\end{figure}

\begin{figure}[H]
\centering
\includegraphics[width=0.21\columnwidth]{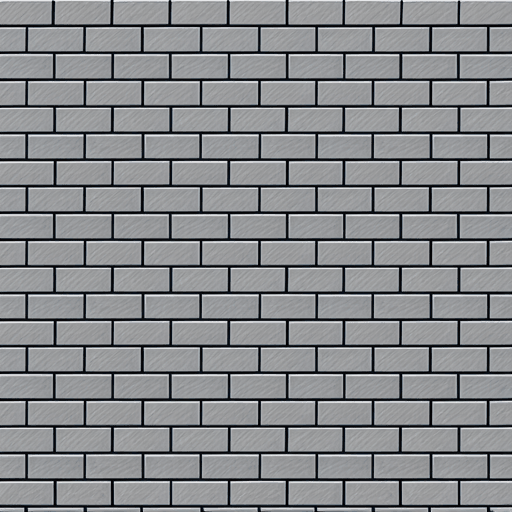}\hspace{0.04\columnwidth}\includegraphics[width=0.21\columnwidth]{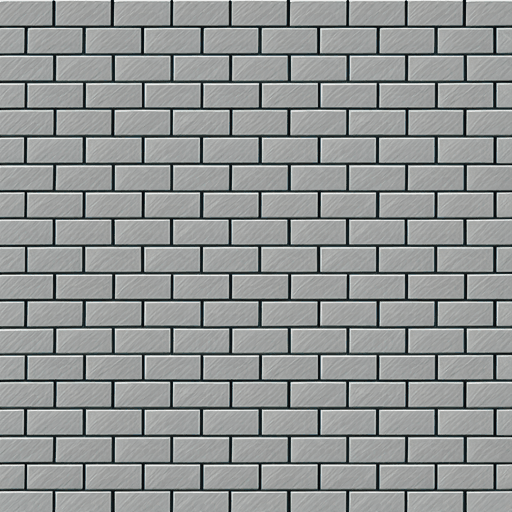}\hspace{0.04\columnwidth}\includegraphics[width=0.21\columnwidth]{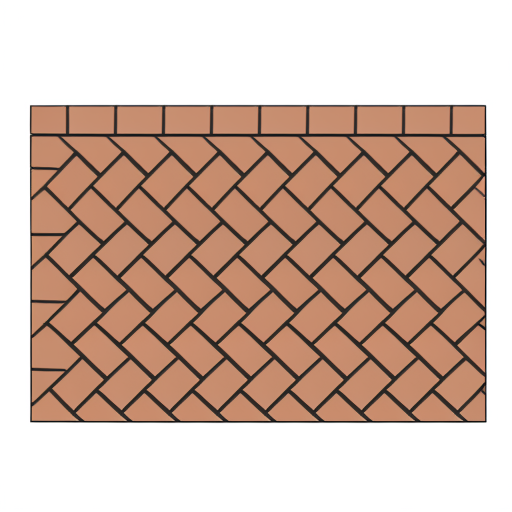}\hspace{0.04\columnwidth}\includegraphics[width=0.21\columnwidth]{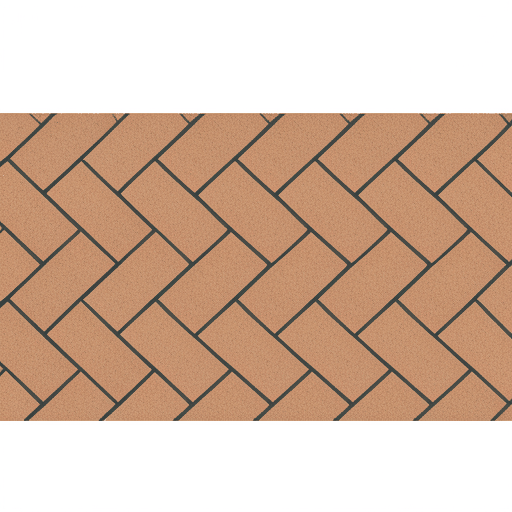}
\caption{Emu3.5, bf16, H100 NVL, stage 10. Worst SSIM pair (prompt 5, grey brick wall): SSIM 0.216; worst LPIPS pair (prompt 7): LPIPS 0.496.}
\label{fig:annex:emubf16}
\end{figure}

\begin{figure}[H]
\centering
\includegraphics[width=0.21\columnwidth]{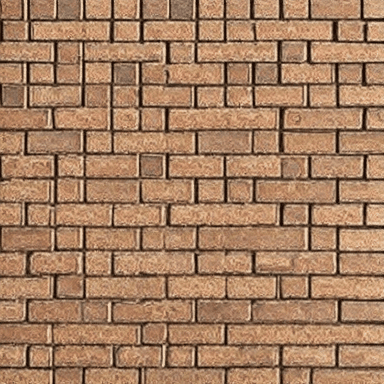}\hspace{0.04\columnwidth}\includegraphics[width=0.21\columnwidth]{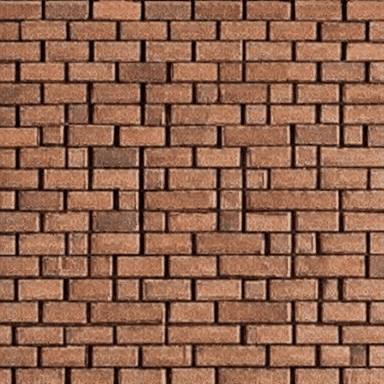}\hspace{0.04\columnwidth}\includegraphics[width=0.21\columnwidth]{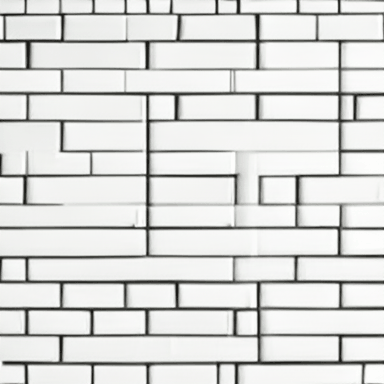}\hspace{0.04\columnwidth}\includegraphics[width=0.21\columnwidth]{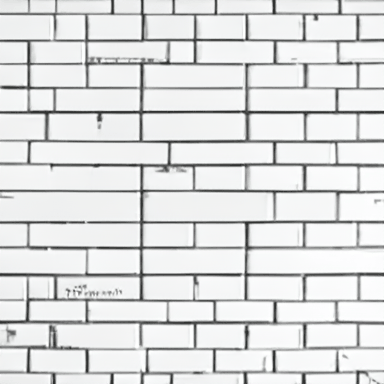}
\caption{Janus-Pro-7B, bf16, RTX 6000 Ada, sampled decoding. Worst SSIM pair (prompt 7, brick wall): SSIM 0.127, LPIPS 0.320; worst LPIPS pair (prompt 6, fine grid): LPIPS 0.474, SSIM 0.259.}
\label{fig:annex:janusint8}
\end{figure}

\section{Execution Details}
\label{app:schedule}
This appendix provides details on the memory and compute time needed to run the experiments on the different hardware platforms for each model.
\refstepcounter{table}\label{tab:resources}
\begin{center}
\footnotesize TABLE \thetable\\[2pt]
\footnotesize Run time and memory usage overview. Times are hours at stages 10, 25, and 100 unless seconds are shown. Memory is the reported peak; sampled nvidia-smi values are lower bounds.\\[3pt]
\scriptsize\setlength{\tabcolsep}{3pt}
\begin{tabular}{@{}p{0.43\columnwidth}p{0.50\columnwidth}@{}}
\hline
Model, precision and GPU & Run time; peak memory \\
\hline
FLUX 3 Klein, bf16, H100-M & 0.32/0.34/1.24 h; 33.4 GB \\
FLUX 3 Klein, int8, H100-M & 0.40/0.44/1.60 h; 27.8 GB \\
FLUX 3 Klein, fp8, H100-M & 0.39/0.42/1.54 h; 27.8 GB \\
HiDream-O1, bf16, H100-M & 0.12/0.14/0.28 h; 18.2 GB \\
HiDream-O1, int8, H100-M & 0.17/0.23/0.55 h; 11.2 GB \\
HiDream-O1, fp8, H100-M & 0.17/0.25/0.63 h; 11.5 GB \\
SDXL, fp16, RTX A6000 & 0.54/1.28/5.03 h; 39.325 GB allocated \\
SDXL, bf16, RTX A6000 & 0.531/1.264/4.74 h; 24.26 GB allocated \\
Qwen-Image-2.1, bf16, H100 NVL & 0.06/0.08/0.17 h; 46.198 GB allocated \\
Qwen-Image-2.1, fp8, H100 NVL & 0.07/0.10/0.24 h; 39.083 GB allocated \\
Emu3.5, bf16, H100 & 0.322/0.762/2.962 h; 92.978 GB sampled \\
Emu3.5, int8, H100 & 0.938/2.271/8.85 h; 81.926 GB sampled \\
\hline
\end{tabular}
\end{center}